\documentclass[runningheads]{llncs}

\usepackage{eccv}

\usepackage{eccvabbrv}

\usepackage{graphicx}
\usepackage{booktabs}

\usepackage[accsupp]{axessibility}  %

\usepackage{multirow}
\usepackage{bm}
\usepackage{kotex}

\usepackage[dvipsnames]{xcolor}
\usepackage{colortbl}
\usepackage{float}
\usepackage{wrapfig}
\usepackage{subcaption}

\newcommand{\red}[1]{{\color{red}{#1}}}

\definecolor{darkergreen}{RGB}{21, 152, 56}
\newcommand{\classtoken}{\texttt{[CLS]}}
\newcommand{\ours}{Temporally Contextualized CLIP}
\newcommand{\Ours}{TC-CLIP}

\definecolor{deemph}{gray}{0.6}

\definecolor{baselinecolor}{gray}{.9}

\newcommand{\notllm}{\cellcolor{NavyBlue!10}}

\usepackage[pagebackref,breaklinks,colorlinks,citecolor=eccvblue]{hyperref}

\usepackage{orcidlink}

\begin{document}

\title{Leveraging Temporal Contextualization for \\ Video Action Recognition} 

\titlerunning{TC-CLIP}

\author{
Minji Kim$^{1\dagger}$ \qquad
Dongyoon Han$^{3}$ \qquad
Taekyung Kim$^{3*}$ \qquad
Bohyung Han$^{1,2*}$
}

\authorrunning{M.~Kim et al.}

\institute{$^{1}$ECE \& $^{2}$IPAI, Seoul National University \qquad
$^{3}$NAVER AI Lab
}

\maketitle
\def\thefootnote{}\footnotetext{
$^{\dagger}$Work done during an internship at NAVER AI Lab.\\
$^{*}$Corresponding authors.}\def\thefootnote{\arabic{footnote}}

\begin{abstract}
We propose a novel framework for video understanding, called \ours~(\Ours), which leverages essential temporal information through global interactions in a spatio-temporal domain within a video.
To be specific, we introduce Temporal Contextualization (TC), a layer-wise temporal information infusion mechanism for videos, which 1) extracts core information from each frame, 2) connects relevant information across frames for the summarization into context tokens, and 3) leverages the context tokens for feature encoding.
Furthermore, the Video-conditional Prompting (VP) module processes context tokens to generate informative prompts in the text modality. 
Extensive experiments in zero-shot, few-shot, base-to-novel, and fully-supervised action recognition validate the effectiveness of our model.
Ablation studies for TC and VP support our design choices.
Our project page with the source code is available at {\small \url{https://github.com/naver-ai/tc-clip}}.
\keywords{Video Action Recognition \and Vision-Language Model}
\end{abstract}
\section{Introduction}\label{sec:intro}

\begin{figure*}[t]
\centering
\includegraphics[width=\linewidth, trim={0 8px 0 4px}, clip]{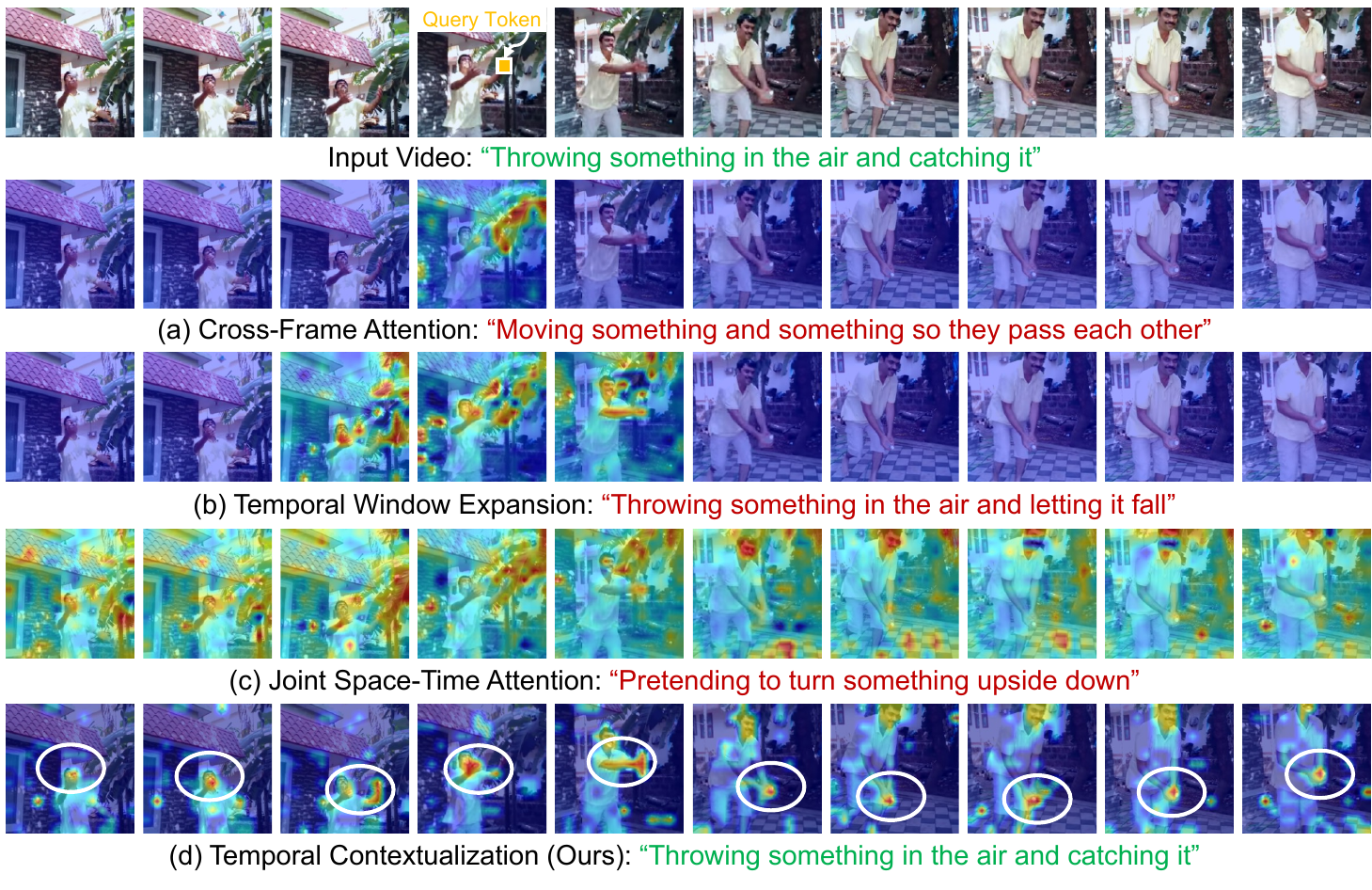}
\vspace{-1em}
\caption{
\textbf{Comparison of attention maps between various temporal modeling approaches.} 
Both (a) and (b) fail to recognize actions in the latter frames, whereas (c) exhibits weak discriminability due to sparse attention on the background.
In contrast, (d) our method successfully focuses on informative regions across all frames, leading to the accurate action recognition result.
}
\label{fig:teaser_attention_visualization}
\end{figure*}

\begin{figure*}[t]
\centering
\includegraphics[width=.98\linewidth]{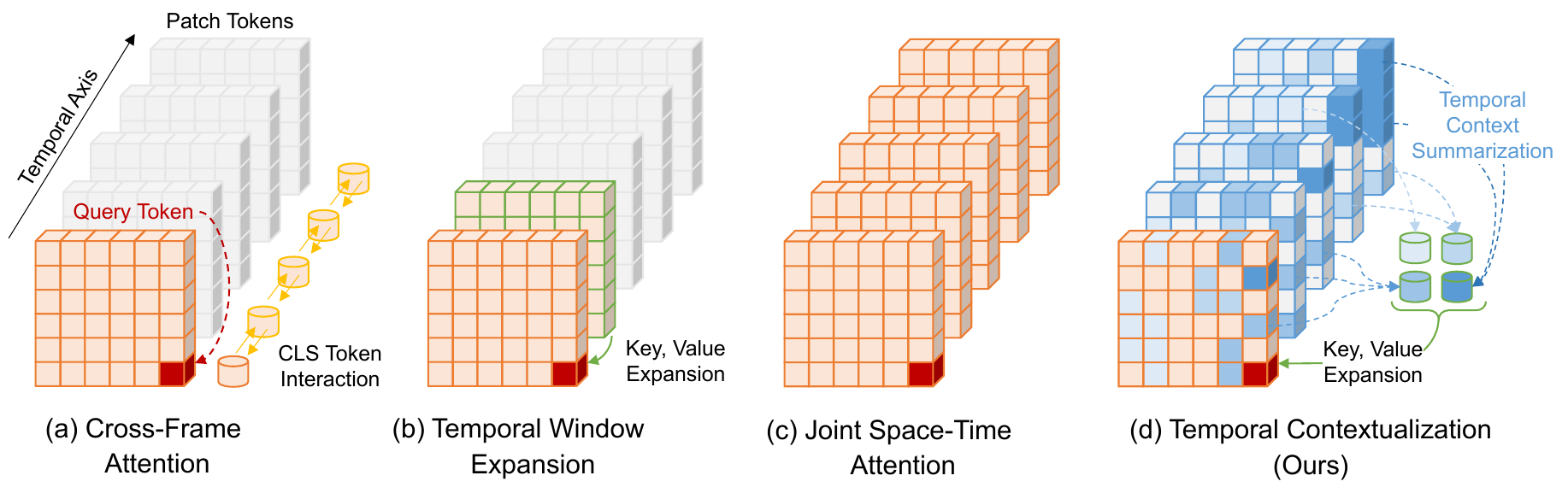}
\vspace{-1em}
\caption{
\textbf{Temporal information learning methods.}
Prior works consider temporal cues during the encoding process via (a) cross-frame attention~\cite{xclip, vitaclip} with \classtoken~token interactions or (b) temporal window expansion~\cite{openvclip} by adding adjacent frame tokens to key-value pairs. However, the former lacks patch-level interactions, while the latter limits the range of temporal interactions. (c) Joint space-time attention allows full interactions across all tokens, but it is costly and suboptimal in practice (see Fig.~\ref{fig:pitfall_joint_attention}.) (d) Unlike prior approaches, our method aggregates pivotal tokens from a broader range yet efficiently for enhanced temporal integration into key-value pairs.
\vspace{-.2em}
}
\label{fig:temporal_attention_comparison}
\end{figure*}
\begin{figure*}[t]
\centering
\includegraphics[width=.98\linewidth, trim={20px 12px 12px 8px}, clip]{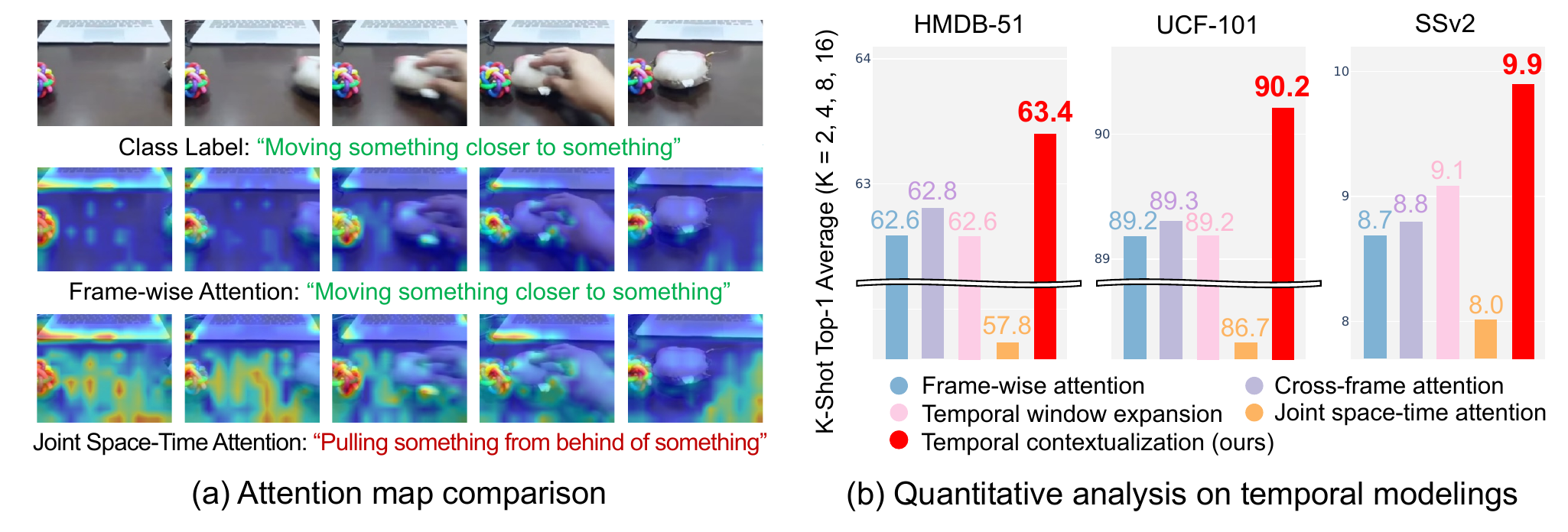}
\caption{\textbf{Pitfall of joint space-time attention.} (a) Extending CLIP's temporal sequence length degrades attention quality, presumably because it was not trained on such long sequences. (b) We compare the action recognition performance in the few-shot setting on diverse datasets. All existing methods fall behind our method.
}
\vspace{-1.5em}
\label{fig:pitfall_joint_attention}
\end{figure*}

Pretrained large-scale Vision-Language Models (VLMs) have shown remarkable generalization capability in video understanding and have emerged as promising tools even for zero-shot or open-vocabulary recognition tasks~\cite{clip, align, florence}.
However, pretraining task-specific models using video-text pairs pose significant challenges, primarily due to substantial computational costs and excessive expense for annotated video-text data~\cite{internvid, videoclip}.
Consequently, recent studies in video understanding~\cite{actionclip, xclip, a5, vitaclip, vificlip, openvclip, ost, froster} have shifted their focus toward employing image-based VLMs such as CLIP~\cite{clip} with fine-tuning for aligning video representations with text embeddings derived from category names.

Despite the success of CLIP in video recognition, existing approaches fail to model temporal information in the video feature learning process, as shown in Fig.~\ref{fig:teaser_attention_visualization}(a)-(b).
This limitation stems from the restrictive token interactions in the temporal axis.
For instance, cross-frame attention approaches~\cite{xclip, vitaclip}, shown in Fig.~\ref{fig:temporal_attention_comparison}(a), attempt to gather temporal information only through class tokens.
Although VCLIP~\cite{openvclip} incorporates patch-level details by bringing keys and values from neighborhood frames in its self-attention operation as in Fig.~\ref{fig:temporal_attention_comparison}(b), its temporal scope is too narrow.
Furthermore, ViFi-CLIP~\cite{vificlip} simply averages frame-wise representations with no inter-frame information exchanges.
Such na\"ive approaches tend to bias the models towards static information in their representation learning (e.g., objects and backgrounds) and hamper learning temporal dynamics (e.g., motion and temporal variations). 
To ensure the global interactions of patch tokens in a spatio-temporal domain, one possible option is to consider every patch token from all frames as a reference during the encoding process as illustrated in Fig.~\ref{fig:temporal_attention_comparison}(c).

Unfortunately, such a straightforward extension for temporally global interactions in VLMs pretrained with short image-text pairs witnesses extrapolation challenges~\cite{press2021train,chen2023extending}; we have observed that a na\"ive extension of sequence length along the temporal axis degrades its discriminability substantially, as shown in Fig.~\ref{fig:pitfall_joint_attention}(a).
The joint space-time attention model spreads attention over many patches and fails to focus on informative tokens to recognize actions, resulting in suboptimal performance compared to the frame-wise attention baseline.
Moreover, this approach suffers from heavy computational overhead due to numerous redundant and similar tokens, which often correspond to background regions.

This paper presents \textbf{Temporally Contextualized CLIP (TC-CLIP)}, a novel paradigm for extending CLIP to videos by encoding holistic video information through advanced temporal analysis.
Specifically, our Temporal Contextualization (TC) pipeline summarizes global action cues into a small set of tokens, called context tokens, for reference during the encoding process.
These context tokens act as additional key-value pairs for attention operations, presumably serving as temporal bridges that convey the video-level context.
Our preliminary study, shown in Fig.~\ref{fig:pitfall_joint_attention}(b), implies that existing methods illustrated in Fig.~\ref{fig:temporal_attention_comparison} offer minimal gains over the frame-wise attention, highlighting the need for enhanced token interactions.

In addition, a Video-conditional Prompting (VP) module generates instance-level textual prompts based on context tokens from the vision encoder.
The VP module comprises cross-attention operations that adopt learnable text prompts as queries and context tokens as keys and values to inject video instance representations into video-conditional textual prompts.
This strategy compensates for the lack of textual semantics in action recognition datasets, where textual descriptions are limited to action class names (e.g., skateboarding, skydiving) without detailed narratives.

To verify the effectiveness and robustness of TC-CLIP, we perform extensive evaluations across diverse video recognition benchmarks.
Quantitative comparisons in zero-shot, few-shot, base-to-novel, and fully-supervised settings show that the proposed approach outperforms the state-of-the-art methods by significant margins.
We also provide an in-depth analysis of our design choices and the impact of each component in our model.
\section{Proposed Method}\label{sec:method}

\subsection{Preliminary}

We first review how CLIP~\cite{clip} is used for video action recognition.
In particular, we discuss the encoding procedure based on the vision and text encoders of CLIP, denoted by $\{ f_{\theta_v}, f_{\theta_c} \}$, to obtain video and text features, $\{ \mathbf{v}, \mathbf{c} \}$.

\vspace{.5em}
\noindent\textbf{Video encoding.}
Suppose that there exists a video $V \in \mathbb{R}^{T \times H \times W \times 3}$ of a spatial resolution $H \times W$ with $T$ sampled frames.
Following the Vision Transformer (ViT) architecture~\cite{vit},
we first divide each frame into $P \times P$ non-overlapping patches and flatten them as a set of vectors $\{ \mathbf{x}_{t,i}  \in \mathbb{R}^{3P^2} \}_{i=1}^N$, 
where $t$ is the frame index, $i$ is the patch index, and $N=HW/P^2$ is the number of patches.
Then, we derive a frame-level token sequence, $\mathbf{z}_t^0$ as follows:
\begin{equation}
     \mathbf{z}_t^0 = [\mathbf{x}_\text{cls},\mathbf{x}_{t,1}\mathbf{W}_\text{emb} , \mathbf{x}_{t,2}\mathbf{W}_\text{emb}, \ldots , \mathbf{x}_{t,N}\mathbf{W}_\text{emb}] + \mathbf{e}_\text{pos},
\end{equation}
where $\mathbf{x}_\text{cls} \in \mathbb{R}^{d_v}$ is a learnable classification embedding named \classtoken~token, $\mathbf{W}_\text{emb} \in \mathbb{R}^{ 3P^2\times d_v}$ is a linear projection matrix, and $\mathbf{e}_\text{pos}$ is the spatial positional embedding.
The CLIP vision encoder, $f_{\theta_v}(\cdot)$, sequentially encodes $\mathbf{z}_t^l$ at each layer $l \in \{1, \ldots, L_v \}$, which is given by
\begin{equation}
    \mathbf{z}_t^l = f_{\theta_v}^l(\mathbf{z}_t^{l-1}). \label{eq:frame_level_rep}
\end{equation}
We project the \classtoken~token of the $t^\text{th}$ frame, denoted by $\mathbf{z}_{t, 0}^{L_v}$, onto a common vision-language latent space using a matrix $\mathbf{W}_{\text{vis}} \in \mathbb{R}^{d_v \times d_{vl}}$, \ie, $\mathbf{v}_t = \mathbf{z}_{t, 0}^{L_v}\mathbf{W}_\text{vis}$.
Finally, the video representation $\mathbf{v}$ is obtained by averaging the per-frame representations $\mathbf{v}_t$ as $\mathbf{v} = \text{AvgPool}([\mathbf{v}_1, ..., \mathbf{v}_T])$.

\vspace{.5em}
\noindent\textbf{Text encoding.}
Given a text description $C$---category name in our problem---for a video, the input of the text encoder, $\textbf{c}^0$, is obtained by tokenizing words in the description and computing their word embedding vectors.
In addition to the embeddings from category names, one can augment a sequence of prompt embeddings $\textbf{p}^0$, which are obtained from either hand-crafted templates such as ``\texttt{a photo of a}'' or learnable prompt vectors.
The CLIP text encoder, $f_{\theta_c}(\cdot)$, sequentially processes a sequence of text embeddings including prompt embeddings, denoted by $[\textbf{p}^0, \textbf{c}^0]$, and computes an intermediate representation at each layer $l \in \{1, \ldots, L_c\}$ as follows:
\begin{equation}
    [\mathbf{p}^{l}, \mathbf{c}^{l}]=f_{\theta_c}^l([\mathbf{p}^{l-1}, \mathbf{c}^{l-1}]),
\end{equation}
where $f_{\theta_c}^l(\cdot)$ denotes the $l^\text{th}$ layer of the CLIP text encoder.
The final text representations $\mathbf{c}$ is obtained by projecting the \texttt{[EOS]} token from the last layer to the vision-language latent space using a matrix $\mathbf{W}_{\text{text}} \in \mathbb{R}^{d_l \times d_{vl}}$, \ie, $\mathbf{c} = \mathbf{c}_{\text{eos}}^{L_c}\mathbf{W}_{\text{text}}$.

\vspace{.5em}
\noindent\textbf{Video-text alignment.}
The similarity between video and text embeddings are formulated as $\mathrm{sim} (\mathbf{v}, \mathbf{c}) = \frac{\langle \mathbf{v}, {\mathbf{c} \rangle}}{\lVert \mathbf{v} \rVert \lVert \mathbf{c} \rVert}$.
During training, the goal is to maximize the similarity if $V$ and $C$ are matched and minimize otherwise.
For inference, the category with the highest similarity is chosen as the prediction.

\subsection{Motivation}

\begin{table*}[t]
\centering
\caption{\textbf{Motivation: What is the proper format for reference tokens?}
We compare 16-shot training results using various types of reference tokens during the frame-level representation encoding process.
Using context tokens consistently improves the baseline model regardless of the choice of the token aggregation function.
\label{tab:motivation}
}
\vspace{-.5em}
\setlength{\tabcolsep}{8pt}
\scalebox{0.8}{
\begin{tabular}{llll}
\toprule
Type of reference tokens & HMDB-51 & UCF-101 & SSv2\\
\midrule
Baseline (No reference tokens) & 67.1 & 93.3 & 12.0\\
\midrule
\classtoken~tokens from all frames~\cite{xclip, vitaclip} & 67.2 (\textcolor{darkergreen}{+0.1}) & 93.2 (\textcolor{red}{$-0.1$}) & 12.3 (\textcolor{darkergreen}{+0.3})\\
Patch tokens from adjacent frames~\cite{openvclip} & 67.8 (\textcolor{darkergreen}{+0.7}) & 93.2 (\textcolor{red}{$-0.1$}) & 12.8 (\textcolor{darkergreen}{+0.8})\\
Patch tokens from all frames& 63.3 (\textcolor{red}{$-3.8$}) & 91.9 (\textcolor{red}{$-1.4$}) & 12.0 (+0.0)\\
\midrule
Context tokens --- K-means~\cite{kmeans} & 68.0 (\textcolor{darkergreen}{+0.9}) & 93.3 (+0.0) & 13.1 (\textcolor{darkergreen}{+1.1})\\
Context tokens --- DPC-KNN~\cite{dpcknn} & 67.9 (\textcolor{darkergreen}{+0.8}) & 94.0 (\textcolor{darkergreen}{+0.7}) & 14.3 (\textcolor{darkergreen}{+2.3})\\
Context tokens --- Bipartite soft matching~\cite{bipartite, tome} & 68.0 (\textcolor{darkergreen}{+0.9}) & 93.8 (\textcolor{darkergreen}{+0.5}) & 14.3 (\textcolor{darkergreen}{+2.3})\\
Context tokens --- Saliency-aware bipartite matching~\cite{vidtldr} & 67.3 (\textcolor{darkergreen}{+0.2}) & 93.7 (\textcolor{darkergreen}{+0.4}) & 13.6 (\textcolor{darkergreen}{+1.6})\\
\bottomrule
\end{tabular}
}
\end{table*}

Despite the successful generalization of CLIP for action recognition, its visual feature encoding process in Eq.~\eqref{eq:frame_level_rep} constrains the model's ability to capture comprehensive spatio-temporal dynamics because it only considers intra-frame token relationships.
This limitation has led previous works to additionally incorporate reference tokens, denoted by $\mathbf{s}$, to encode the $t^\text{th}$ frame tokens $\mathbf{z}_t$ as
\begin{equation}
    \mathbf{z}_t^l = f_{\theta_v}^l(\mathbf{z}_t^{l-1}, \mathbf{s}^{l-1}).\label{eq:frame_level_rep_extend}
\end{equation}
However, their reference token designs are still limited due to insufficient spatio-temporal interaction range.
For instance, cross-frame attention (Fig.~\ref{fig:temporal_attention_comparison}(a))~\cite{xclip, vitaclip} utilizes learnable global embedding vectors, \eg, \classtoken~tokens, from all frames to define the reference token as $\mathbf{s} = [\mathbf{z}_{1, 0}, ..., \mathbf{z}_{T, 0}]$, and temporal window expansion (Fig.~\ref{fig:temporal_attention_comparison}(b))~\cite{openvclip}, on the other hand, integrates neighboring frame patch tokens for $\mathbf{s} = [\mathbf{z}_{t-1}, \mathbf{z}_{t+1}]$.
Note that the former lacks patch-level details whereas the latter captures temporal information only within a local range.
Although incorporating all patch tokens from a whole video as a reference (Fig.~\ref{fig:temporal_attention_comparison}(c)), $\mathbf{s} = [\mathbf{z}_{1}, ..., \mathbf{z}_{T}]$, might be conceptually reasonable, it is not practical due to the excessive number of tokens.
Furthermore, this approach conflicts with the properties of CLIP pretrained on short image-text pairs and significantly degrades attention quality.

To this end, we compute a reference, $\mathbf{s} = \phi([\mathbf{z}_{1}, ..., \mathbf{z}_{T}])$, using a small set of context tokens that summarize a whole input video, where $\phi(\cdot)$ is a token aggregation function.
This approach delivers the essential temporal information for the feature encoding while maintaining CLIP's effective sequence length.
Our preliminary study, presented in Table~\ref{tab:motivation}, shows that using the proposed context tokens as a reference consistently outperforms the frame-wise attention baseline, whereas all other types of reference tokens only yield marginal gains or even suffer from performance degradation.

\subsection{Temporal Contextualization (TC)\label{sec:method_tc}}

\begin{figure*}[t]
\centering
\includegraphics[width=\linewidth, trim={0 10px 0 10px}, clip]{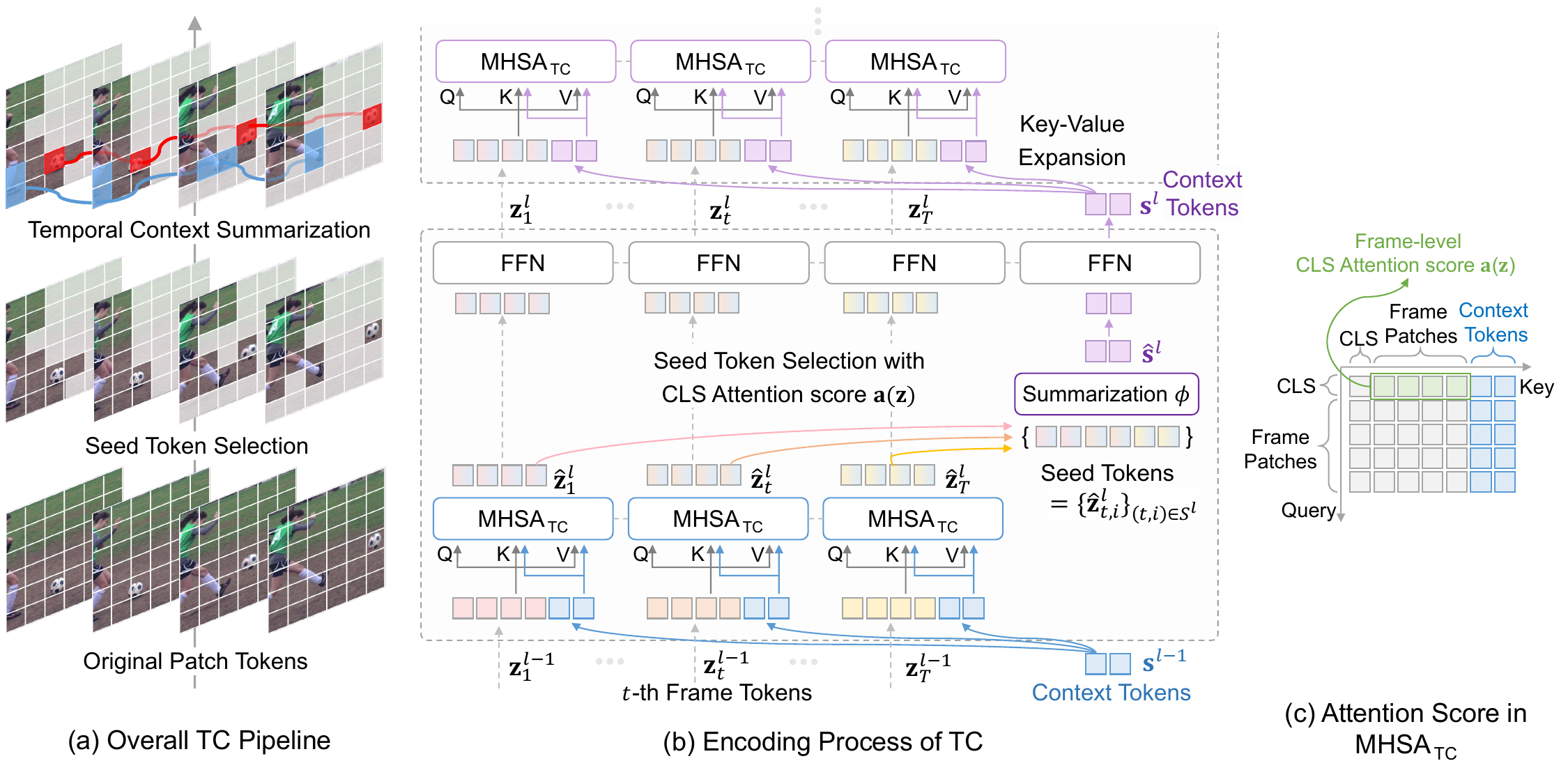}
\caption{
\textbf{Overview of Temporal Contextualization (TC).}
We first collect informative tokens from each frame and then aggregate relevant seed tokens to obtain context tokens.
They are used as key-value pairs for the self-attention in the next layer.
}
\label{fig:temporal_information_propagation}
\end{figure*}

Based on our motivation, we propose Temporal Contextualization (TC), which consists of three steps: 1) informative token selection in each frame, 2) context summarization across spatio-temporal dimensions, and 3) context infusion to all tokens in the subsequent layer.
Fig.~\ref{fig:temporal_information_propagation} illustrates an overview of TC.

\vspace{.5em}
\noindent\textbf{Informative token selection.}
Due to the many redundant tokens in a video, using all patches may be suboptimal for extracting desired temporal information.
We only select the informative seed tokens based on each frame's attention scores obtained from self-attention operations.
Specifically, given patch tokens $\{\mathbf{z}_{t,i}\}_{i=1}^N$ in the $t^\text{th}$ frame, a set of attention scores $\{\mathbf{a}(\mathbf{z}_{t,i})\}_{i=1}^N$ is driven from the attentiveness of the \classtoken~token with respect to other tokens, which is given by
\begin{equation}
    \mathbf{a}(\mathbf{z}_t) = \text{Softmax}\Big(\frac{\mathbf{q}_{\text{cls}} \mathbf{K}_{\mathbf{z}_t}^\mathsf{T}}{\sqrt{d}}\Big),
\end{equation}
where both the query $\mathbf{q}_\text{cls}=\mathbf{z}_\text{t,0} \mathbf{W}_q \in \mathbb{R}^{d}$ and keys $\mathbf{K}_{\mathbf{z}_t}=\mathbf{z}_t\mathbf{W}_k \in \mathbb{R}^{(N+1) \times d}$ are given by linear projections of input $\mathbf{z}_t \in \mathbb{R}^{(N+1) \times d}$.
In practice, our model yields multiple \classtoken~attention scores from multi-head self-attention (MHSA) operations and computes the average of the attention scores from all heads, \ie, $\bar{\mathbf{a}}_{t,i} = \sum_{h=1}^H \mathbf{a}_{t,i}^h / H$, where $ \mathbf{a}_{t,i}^h=\mathbf{a}^h(\mathbf{z}_{t,i})$ is the attention score for the $i^\text{th}$ patch $\mathbf{z}_{t,i}$ in the $t^\text{th}$ frame and $H$ is the number of heads.
Finally, we identify a set of seed token indices for the $t^\text{th}$ frame, $\mathcal{S}_t$, by selecting $n_s$ elements with the highest attention scores, where $n_s$ is controlled by a hyperparameter $\alpha = n_s/N$.

\vspace{.5em}
\noindent\textbf{Temporal context summarization.}
We describe how to connect the seed tokens derived from individual frames based on their relevance and identify a collection of context tokens.
We first collect the seed tokens from all frames as $\{\hat{\mathbf{z}}_{t,i}\}_{(t,i) \in \mathcal{S}}$, where $\mathcal{S} = \{(t,i)|i\in\mathcal{S}_t, t=1,\ldots,T\}$ is a set of seed token indices from all frames and $\hat{\textbf{z}}_{t,i}$ indicates an interim token encoded from $\textbf{z}_{t,i}$ via the self-attention operation.
Then we perform their spatio-temporal summarization by clustering and merging all the seed tokens as
\begin{equation}
    \hat{\mathbf{s}} = \phi\big(\{\hat{\mathbf{z}}_{t,i}\}_{(t,i) \in \mathcal{S}}\big),
\end{equation}
where $\hat{\mathbf{s}} \in \mathbb{R}^{k \times D}$ denotes a collection of the summarized tokens, which we call context tokens, and $\phi$ is a token aggregation function.
While diverse token aggregation techniques are valid for TC (See Table~\ref{tab:abl_aggregation}), we adopt bipartite soft matching~\cite{bipartite, tome} by default.
Subsequently, the context tokens $\hat{\mathbf{s}}$ are fed into a feed-forward network (FFN).

\vspace{.5em}
\noindent\textbf{Temporal context infusion.}
Finally, we infuse the summarized context to all patch tokens by modifying the self-attention function. The keys and values of self-attention in every frame are expanded to include context tokens as follows:
\begin{equation}\label{eq:attention_tip}
    \text{Attention}_{\text{TC}}(\mathbf{z}_t, \mathbf{s}) = \text{Softmax}\Big(\frac{\mathbf{Q}_{\mathbf{z}_t} \big[\mathbf{K}_{\mathbf{z}_t}|\mathbf{K}_{\mathbf{s}}\big]^\mathsf{T}}{\sqrt{d}} + \mathbf{B} \Big) \big[\mathbf{V}_{\mathbf{z}_t}|\mathbf{V}_{\mathbf{s}}\big],
\end{equation}
where $\mathbf{K}_\mathbf{s} = \mathbf{s}\mathbf{W}_k$ and $\mathbf{V}_{\mathbf{s}} = \mathbf{s}\mathbf{W}_v$ are linear projections of the context tokens $\mathbf{s} \in \mathbb{R}^{k \times d}$.
Here, $\mathbf{B} \in \mathbb{R}^{(N+1) \times (N+k+1)}$ is a bias matrix that distinguishes between frame-level local information and video-level global information in the expanded key matrix as follows:
\begin{equation}\label{eq:learnable_bias}
\mathbf{B}_{ij} = 
\begin{cases} 
b_{\text{local}} & \text{if } j \leq N + 1 \\
b_{\text{global}} & \text{otherwise},
\end{cases}
\end{equation}
where $b_{\text{local}}$ and $b_{\text{global}}$ are learnable parameters and defined for multiple heads at each layer.
We build our TC pipeline in a layer-wise manner, and thus the encoding process of each layer is expressed as
\begin{align}
    \hat{\mathbf{z}}_t^l &= 
    \begin{cases} 
        \text{MHSA}(\text{LN}(\mathbf{z}_t^{l-1})) + \mathbf{z}_t^{l-1} & \text{if } l=1\\
        \text{MHSA}_\text{TC}(\text{LN}(\mathbf{z}_t^{l-1}), \text{LN}(\mathbf{s}^{l-1})) + \mathbf{z}_t^{l-1} & \text{otherwise},
    \end{cases}\\
    \mathbf{z}_t^l & =\text{FFN}(\text{LN}(\hat{\mathbf{z}}_t^l)) + \hat{\mathbf{z}}_t^l,\\
    \mathbf{s}^l &= \text{FFN}(\text{LN}(\hat{\mathbf{s}}^l)) + \hat{\mathbf{s}}^l,
\end{align}
where $\text{MHSA}_\text{TC}(\cdot, \cdot)$ denotes the MHSA operation based on Eq.~\eqref{eq:attention_tip} and $\text{LN}(\cdot)$ stands for the layer normalization function.

\subsection{Video-conditional Prompting (VP) \label{sec:method_vp}}

\begin{figure*}[t]
\centering
\includegraphics[width=\linewidth, trim={0 0 0 8px}, clip]{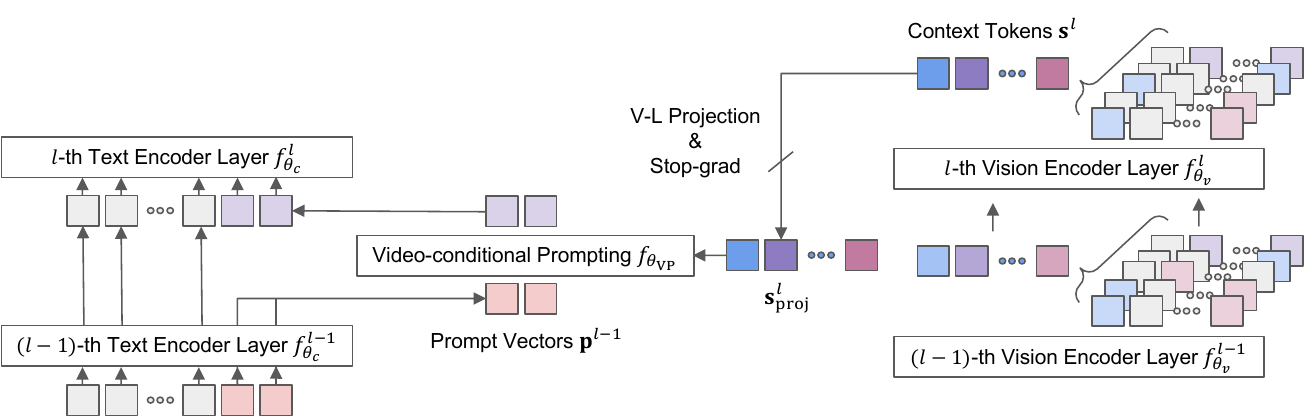}
\caption{
\textbf{Video-conditional Prompting (VP) module.}
Video information from the context tokens is injected into the text prompt vectors using a cross-attention mechanism, generating instance-level prompts that make up for the lack of textual semantics.
}
\label{fig:modality_interaction}
\end{figure*}

The Video-conditional Prompting (VP) module further leverages the comprehensive video information derived from the visual domain for text encoding.
We apply a cross-attention between prompt vectors and context tokens to enrich the information in the prompt vectors as illustrated in Fig.~\ref{fig:modality_interaction}. 
Let $\textbf{c}^{l-1}$ and $\textbf{p}^{l-1}$ be class name tokens and learnable prompt vectors from the $(l-1)^\text{th}$ layer of the text encoder, respectively.
We derive temporally contextualized prompt vectors $\Tilde{\mathbf{p}}^{l-1}$ by passing the layer-normalized prompt tokens and context tokens through a cross-attention layer and an FFN layer as follows:
\begin{align}
    \mathbf{s}_{\text{proj}}^l &= \text{SG}(\mathbf{s}^l\mathbf{W}_{\text{vis}}),\label{eq:vp_proj} \\
    \hat{\mathbf{p}}^{l-1} &= \text{MHCA}(\text{LN}_p(\mathbf{p}^{l-1}), \text{LN}_s(\mathbf{s}_{\text{proj}}^l)) + \mathbf{p}^{l-1}, \label{eq:vp_mhca} \\
    \Tilde{\mathbf{p}}^{l-1} &= \text{FFN}(\text{LN}(\hat{\mathbf{p}}^{l-1})) + \hat{\mathbf{p}}^{l-1},\label{eq:vp_ffn}
\end{align}
where $\text{SG}(\cdot)$ is a stop-gradient function, 
$\mathbf{W}_{\text{vis}}$ is a weight matrix of CLIP to linearly project vision representations onto a common vision-language latent space, 
and $\text{MHCA}(\cdot, \cdot)$ is a multi-head cross-attention operation for interactions across modalities, accepting text prompt vectors as queries and vision features as keys and values.
The VP module $f_{\theta_{\text{VP}}}(\cdot, \cdot)$ is defined by a composition of Eq.~\eqref{eq:vp_mhca} and Eq.~\eqref{eq:vp_ffn}, and executed before the last layer of text encoder $f_{\theta_c}$.
Finally, the new formulation of our encoding process in the text modality is given by
\begin{equation}
[\mathbf{p}^l, \mathbf{c}^l] =
    \begin{cases}
    f_{\theta_c}^l([f_{\theta_{\text{VP}}}(\mathbf{p}^{l-1},\mathbf{s}_{\text{proj}}^l), \mathbf{c}^{l-1}])  & \text{if}~l=L_c \\
     f_{\theta_c}^l([\mathbf{p}^{l-1}, \mathbf{c}^{l-1}]) & \text{otherwise.}
    \end{cases}
\end{equation}

\subsection{Training Objective}
TC-CLIP learns to maximize the similarity of video representations $\mathbf{v}$ and text representations $\mathbf{c}$ for matching pairs in a mini-batch via the cross-entropy loss as 
$\mathcal{L} = -\sum_{i} \log\frac{\text{exp}(\text{sim}(\mathbf{v}_i,\mathbf{c}_i)/\tau)}{\sum_{j}\text{exp}(\text{sim}(\mathbf{v}_i, \mathbf{c}_{j})/\tau)},$
where $\tau$ is a learnable temperature parameter.
Our model is fully fine-tuned in an end-to-end manner.

\section{Experiments}\label{sec:experiment}

\begin{table*}[t]
\centering
\caption{
\textbf{Comparison with state-of-the-arts on zero-shot action recognition.} All the models are trained on Kinetics-400 and directly evaluated on other datasets.
WE indicates the weight-space ensemble between the fine-tuned model and CLIP, adopted for all applicable models for fair comparisons.
\textdagger~denotes results reproduced using our implementation.
The best results are in \textbf{bold-faced} numbers, and the second-best ones are \underline{underlined}.
Our results using the original and LLM-rephrased category names are highlighted in \colorbox{NavyBlue!10}{blue} and \colorbox{Purple!10}{purple}, respectively.
}
\label{p1_zeroshot}
\vspace{-.5em}
\hspace{-3mm}
\setlength{\tabcolsep}{7pt}
\scalebox{0.8}{
\begin{tabular}{lcccccc} \toprule
Method & WE  & HMDB-51 & UCF-101 & K600 (Top-1) & K600 (Top-5) & All (Top-1)\\ 
\midrule
Vanilla CLIP~\cite{clip}                &        & 40.8 $\pm$ 0.3                & 63.2 $\pm$ 0.2                & 59.8 $\pm$ 0.3                & 83.5 $\pm$ 0.2                & 54.6 \\
ActionCLIP~\cite{actionclip}$^\dagger$         &        & 49.1 $\pm$ 0.4                & 68.0 $\pm$ 0.9                & 56.1 $\pm$ 0.9                & 83.2 $\pm$ 0.2                & 57.7 \\
A5~\cite{a5}                            &        & 44.3 $\pm$ 2.2                & 69.3 $\pm$ 4.2                & 55.8 $\pm$ 0.7                & 81.4 $\pm$ 0.3                & 56.5 \\
X-CLIP~\cite{xclip}                     &        & 44.6 $\pm$ 5.2                & 72.0 $\pm$ 2.3                & 65.2 $\pm$ 0.4                & 86.1 $\pm$ 0.8                & 60.6  \\
Vita-CLIP~\cite{vitaclip}               &        & 48.6 $\pm$ 0.6                & 75.0 $\pm$ 0.6                & 67.4 $\pm$ 0.5                & -                             & 63.7 \\
ViFi-CLIP~\cite{vificlip}$^\dagger$            &        & \underline{52.3} $\pm$ 0.2    & \underline{78.9} $\pm$ 1.1    & \underline{70.7} $\pm$ 0.8    & \underline{92.1} $\pm$ 0.3    & \underline{67.3} \\
\rowcolor{NavyBlue!10}
\Ours~(Ours)                            &        & \textbf{53.7} $\pm$ 0.7       & \textbf{80.4} $\pm$ 0.9       & \textbf{72.7} $\pm$ 0.5       & \textbf{93.2} $\pm$ 0.2       & \textbf{68.9} \\ 
\midrule
ActionCLIP~\cite{actionclip}$^\dagger$ &  \checkmark  & 51.9 $\pm$ 0.5                & 74.2 $\pm$ 1.0                & 67.5 $\pm$ 1.2                & 90.7 $\pm$ 0.1                & 64.5 \\
ViFi-CLIP~\cite{vificlip}$^\dagger$    &  \checkmark  & 52.2 $\pm$ 0.7                & 81.0 $\pm$ 0.9                & \underline{73.9} $\pm$ 0.5    & \underline{93.3} $\pm$ 0.3    & 69.0 \\
Open-VCLIP~\cite{openvclip}     &  \checkmark  & \underline{53.9} $\pm$ 1.2    & \textbf{83.4} $\pm$ 1.2       & 73.0 $\pm$ 0.8                & 93.2 $\pm$ 0.1                & \underline{70.1} \\
\rowcolor{NavyBlue!10}
\Ours~(Ours)                    &  \checkmark  & \textbf{54.2} $\pm$ 0.7       & \underline{82.9} $\pm$ 0.6    & \textbf{75.8} $\pm$ 0.5       & \textbf{94.4} $\pm$ 0.2       & \textbf{71.0} \\ 
\midrule
\multicolumn{6}{l}{\textit{Using LLM-based text augmentation}} \vspace{1px} \\
MAXI~\cite{maxi}                &  \checkmark  & 52.3 $\pm$ 0.7                & 78.2 $\pm$ 0.8                & 71.5 $\pm$ 0.8                & 92.5 $\pm$ 0.4                & 67.3\\
OST~\cite{ost}                  &  \checkmark  & \underline{55.9} $\pm$ 1.2    & 79.7 $\pm$ 1.1                & \underline{75.1} $\pm$ 0.6    & \underline{94.6} $\pm$ 0.2    & 70.2 \\
FROSTER~\cite{froster}          &  \checkmark  & 54.8 $\pm$ 1.3                & \underline{84.8} $\pm$ 1.1    & 74.8 $\pm$ 0.9                & -                             & \underline{71.5} \\
\rowcolor{Purple!10}
\Ours~(Ours)                    &  \checkmark  & \textbf{56.0} $\pm$ 0.3       & \textbf{85.4} $\pm$ 0.8       & \textbf{78.1} $\pm$ 1.0       & \textbf{95.7} $\pm$ 0.3       & \textbf{73.2}   \\
\bottomrule
\end{tabular}
}
\end{table*}
\begin{table*}[t]
\centering
\caption{\textbf{Comparison with state-of-the-arts on few-shot action recognition.} All the models are directly fine-tuned from CLIP. 
Our results using the original and LLM-rephrased category names are highlighted in \colorbox{NavyBlue!10}{blue} and \colorbox{Purple!10}{purple}, respectively.
}
\label{p1_few_shot}
\vspace{-1em}
\hspace{-3mm}
\setlength{\tabcolsep}{2.8pt}
\scalebox{0.8}{
\begin{tabular}{l cccc|cccc|cccc|c}
\toprule
&\multicolumn{4}{c}{HMDB-51} & \multicolumn{4}{c}{UCF-101} & \multicolumn{4}{c}{SSv2} & All\\
\cmidrule(lr){2-14}
Method & $K$=$2$ & $K$=$4$ & $K$=$8$ & $K$=$16$ & $K$=$2$ & $K$=$4$ & $K$=$8$ & $K$=$16$ & $K$=$2$ & $K$=$4$ & $K$=$8$ & $K$=$16$ & ~Avg.~\\
\midrule
Vanilla CLIP~\cite{clip}      & 41.9	& 41.9	& 41.9	& 41.9	& 63.6	&63.6	&63.6	&63.6	&2.7	&2.7	&2.7	&2.7    & 36.1 \\
ActionCLIP~\cite{actionclip}  & 47.5	& 57.9	& 57.3	& 59.1	& 70.6	&71.5	&73.0	&91.4	&4.1	&5.8	&8.4	&11.1   & 46.5 \\
A5~\cite{a5}                  & 39.7	& 50.7	& 56.0	& 62.4	& 71.4	&79.9	&85.7	&89.9	&4.4	& 5.1	& 6.1	& 9.7   & 46.8 \\
X-CLIP~\cite{xclip}           & 53.0	& 57.3	& 62.8	& 64.0	& 76.4	&83.4	&88.3	&91.4	&3.9	&4.5	&6.8	&10.0   & 50.2 \\
ViFi-CLIP~\cite{vificlip}     & \underline{57.2}	& \textbf{62.7}	& \underline{64.5}	& \underline{66.8}	& \underline{80.7}	& \underline{85.1}	& \underline{90.0}	& \underline{92.7}	& \underline{6.2}	& \underline{7.4}	& \underline{8.5}	& \underline{12.4}   & \underline{52.9} \\ 
\rowcolor{NavyBlue!10}
\Ours~(Ours)                  & \textbf{57.3} & \underline{62.3}	& \textbf{67.3}	& \textbf{68.6}	& \textbf{85.9}	& \textbf{89.9}	& \textbf{92.5}	& \textbf{94.6}	&\textbf{7.3}	&\textbf{8.6}	&\textbf{9.3}	&\textbf{14.0}& \textbf{54.8}\\ 
\midrule
\multicolumn{14}{l}{\textit{Using LLM-based text augmentation}} \vspace{1px} \\
OST~\cite{ost}                             & \textbf{59.1}  & \underline{62.9}	& \underline{64.9}	& \underline{68.2}	& \underline{82.5}	& \underline{87.5}	& \underline{91.7}	& \underline{93.9}	& \underline{7.0}	& \underline{7.7}	& \underline{8.9}	& \underline{12.2}  & \underline{53.9} \\ 
\rowcolor{Purple!10}
\Ours~(Ours) & \underline{58.6} & \textbf{63.3} & \textbf{65.5} & \textbf{68.8}   & \textbf{86.8} & \textbf{90.1} & \textbf{92.0} & \textbf{94.3}   & \notllm \textbf{7.3}	& \notllm \textbf{8.6}	& \notllm \textbf{9.3}	& \notllm \textbf{14.0}   & \textbf{54.9} \\
\bottomrule                             
\end{tabular}
}
\vspace{-.5em}
\end{table*}

\begin{table*}[t!]
\centering
\caption{\textbf{Comparison with state-of-the-arts on base-to-novel generalization.} 
All the models are directly fine-tuned from CLIP.
\textdagger~results are taken from~\cite{froster}.
}
\label{tab:base2novel}
\vspace{-1em}
\hspace{-3mm}
\setlength{\tabcolsep}{1.5pt}
\scalebox{0.8}{
\begin{tabular}[t]{lccc|ccc|ccc|ccc|ccc}
\toprule
&      \multicolumn{3}{c}{K-400}    &\multicolumn{3}{c}{HMDB-51}  &\multicolumn{3}{c}{UCF-101}    & \multicolumn{3}{c}{SSv2} & \multicolumn{3}{c}{All (Avg.)} \\
\cmidrule(lr){2-16}
Method  & Base & Novel & HM          & Base & Novel & HM         & Base & Novel & HM         & Base & Novel & HM   & Base & Novel & HM   \\
\midrule
Vanilla CLIP~\cite{clip}            & 62.3 & 53.4 & 57.5        & 53.3 &  46.8 & 49.8       & 78.5 & 63.6 & 70.3       & 4.9 & 5.3 & 5.1 & 49.8 & 42.3 & 45.7 \\ 
ActionCLIP~\cite{actionclip}        & 61.0 & 46.2 & 52.6         & 69.1 & 37.3 & 48.5        & 90.1 & 58.1 & 70.7       &  13.3 &  10.1 &  11.5  &  58.5 &  37.9 &  46.0 \\
A5~\cite{a5}                        & 69.7 & 37.6 & 48.8        & 46.2 & 16.0 & 23.8        &  90.5 & 40.4 & 55.8       & 8.3 & 5.3 & 6.4  &  53.7 &  24.8 &  33.9\\
X-CLIP~\cite{xclip}                 & 74.1 &  56.4 &  64.0        &  69.4 & 45.5 &  55.0     & 89.9 &  58.9 &  71.2       & 8.5 & 6.6 & 7.4 &  60.5 &  41.9 &  49.5\\
ViFi-CLIP~\cite{vificlip}           & 76.4 & 61.1 & 67.9        & \textbf{73.8} & \underline{53.3} & \underline{61.9}        &  92.9 & 67.7 & 78.3       & \underline{16.2} & \underline{12.1} & \underline{13.9} &  \underline{64.8} &  \underline{48.6} &  55.5\\
Open-VCLIP~\cite{openvclip}$^\dagger$      & \underline{76.5} & \underline{62.6} & \underline{68.9}        & 70.3 & 50.4 & 58.9        &  \underline{94.8} & \underline{77.5} & \underline{85.3}       & 16.0 & 11.0 & 13.0  &  64.4 &  50.4 &  \underline{56.5}\\
\rowcolor{NavyBlue!10}
\Ours~(Ours)  & \textbf{78.9} & \textbf{63.6} & \textbf{70.4}        & \underline{73.3} & \textbf{54.1} & \textbf{62.2}        & \textbf{95.5} & \textbf{78.0} & \textbf{85.9}       & \textbf{17.5} & \textbf{13.4} & \textbf{15.2}  &  \textbf{66.3} &  \textbf{52.3} &  \textbf{58.5}\\ 
\midrule
\multicolumn{13}{l}{\textit{Using LLM-based text augmentation}} \vspace{1px} \\
FROSTER~\cite{froster}        & \underline{77.8} & \underline{64.3} & \underline{70.4}        & \textbf{74.1} & \underline{58.0} & \underline{65.1}        &  \underline{95.3} & \underline{80.0} & \underline{87.0}       & \textbf{18.3} & \underline{12.2} & \underline{14.6} & \textbf{66.4} & \underline{53.6} & \underline{59.3} \\
\rowcolor{Purple!10}
\Ours~(Ours)  & \textbf{79.1}  & \textbf{65.4} & \textbf{71.6}        & \underline{73.3}  & \textbf{59.1} & \textbf{65.5}      & \textbf{95.4} & \textbf{81.6} & \textbf{88.0}  & \notllm \underline{17.5} & \notllm \textbf{13.4} & \notllm \textbf{15.2} & \underline{66.3} & \textbf{54.9} & \textbf{60.1} \\
\bottomrule
\end{tabular}
}
\vspace{-1em}
\end{table*}
\begin{figure}[t]
\begin{minipage}{1.0\textwidth}
\vspace{-1em}
\begin{minipage}[t]{0.4\textwidth}
\captionof{table}
{\textbf{Fully-supervised action recognition results on Kinetics-400.}
Views means (temporal clips)\,$\times$\,(spatial crops), and F denotes number of frames.
\label{p1_fullysupervised}
}
\centering
\tabcolsep=2.2pt
\vspace{.5em}
\hspace{-3mm}
\scalebox{0.8}{
\begin{tabular}{lcccc}
\toprule
Method & Top-1 &  Top-5 & F & Views \\ 
\midrule
ActionCLIP~\cite{actionclip}       &   83.8  & 96.2 & 32      &  10\,$\times$\,3 \\
X-CLIP~\cite{xclip}                & \underline{84.7} &  \underline{96.8} & 16 & 4\,$\times$\,3 \\
Vita-CLIP~\cite{vitaclip}          & 82.9 &  96.3 & 16 & 4\,$\times$\,3 \\
ViFi-CLIP~\cite{vificlip}   	   &   83.9  &  96.3 &  16  &  4\,$\times$\,3 \\ 
OST~\cite{ost}   	               &   83.2  & -  &  16  & 1\,$\times$\,1 \\
\rowcolor{NavyBlue!10}
\Ours~(Ours)   	                   &   \textbf{85.2}  &  \textbf{96.9}  &  16  &  4\,$\times$\,3  \\ 
\bottomrule
\end{tabular}%
 }
\end{minipage}
\noindent\hfill
\begin{minipage}[t]{0.58\textwidth}
\captionof{table}
{
\textbf{Computational costs} with the average top-1 accuracies of all protocols. The Throughput per view (TP) is measured on a single A6000 GPU. \S~denotes that TC is partly applied to the 4$^\text{th}$, 8$^\text{th}$, and 12$^\text{th}$ layers of the vision encoder. 
}
\centering
\tabcolsep=1pt
\vspace{.5em}
\hspace{-2mm}
\scalebox{0.8}{
\begin{tabular}{lccccccc} 
\toprule
Method & Zero & Few & B2N & Full & Params & GFLOPs & TP\\ 
\midrule
ActionCLIP~\cite{actionclip} & 64.5 & 46.5 & 46.0 & 83.8 & 143.7M & 567 & 20\\
X-CLIP~\cite{xclip} & 60.6 & 50.2 & 49.5 & 84.7 & 169.7M& \underline{288} & \underline{36}\\
Vita-CLIP~\cite{vitaclip} & 63.7 & - & - & 82.9 & 161.8M & 307 & 30 \\
ViFi-CLIP~\cite{vificlip} & 69.0 & 52.9 & 55.5 & 83.9 & \textbf{124.3M} & \textbf{285} & \textbf{38}\\
Open-VCLIP~\cite{openvclip} & 70.1 & - & 56.5 & - & \textbf{124.3M} & 308 & 29\\
\rowcolor{NavyBlue!10}
\Ours~(Ours) & \textbf{71.0} & \textbf{54.8} & \underline{58.5} & \textbf{85.2} & \underline{127.5M} & 304 & 24\\
\rowcolor{NavyBlue!10}
\Ours~(Ours)$^\S$ & \underline{70.7} & \underline{54.4} & \textbf{58.6} & \underline{84.9} & \underline{127.5M} & 291 & 34\\
\bottomrule
\end{tabular}
}
\label{tab:computation_cost_comparison}
\end{minipage}
\end{minipage}
\end{figure}

We conduct experiments on 5 video benchmarks: Kinetics-400~\cite{k400} \& 600~\cite{k600}, HMDB-51~\cite{hmdb51}, UCF-101~\cite{ucf101}, and Something-Something v2 (SSv2)~\cite{ssv2}.
Following~\cite{vificlip}, our evaluation protocols include zero-shot, few-shot, base-to-novel generalization, and fully-supervised action recognition tasks.
We adopt CLIP with ViT-B/16 for all experiments and our baseline is ViFi-CLIP~\cite{vificlip}.
All models are trained using 4 NVIDIA Tesla V100 GPUs.
More details are in the appendix.

\subsection{Quantitative Comparison}

We mainly compare our method with CLIP-based video recognition models: Vanilla CLIP~\cite{clip}, ActionCLIP~\cite{actionclip}, A5~\cite{a5}, X-CLIP~\cite{xclip}, Vita-CLIP~\cite{vitaclip}, ViFi-CLIP~\cite{vificlip}, Open-VCLIP~\cite{openvclip}, OST~\cite{ost}, and FROSTER~\cite{froster}.
For the fair comparisons with approaches based on Large Language Model (LLM) with text augmentation~\cite{maxi, ost, froster}, we produce two versions of our results: one using the original action category names (colored in \colorbox{NavyBlue!10}{blue}) and the other adopting the LLM-rephrased category names obtained from FROSTER~\cite{froster} (colored in \colorbox{Purple!10}{purple}).
Note that experiments on the SSv2 dataset do not involve LLM-rephrasing.

\vspace{.5em}
\noindent\textbf{Zero-shot action recognition.}
Table~\ref{p1_zeroshot} exhibits the zero-shot generalization ability of several models, where they are trained on K-400 and then directly evaluated on individual datasets.
For fair comparisons with recent models~\cite{openvclip, maxi, ost, froster}, we employ weight-space ensembling (WE) for all applicable models except those freezing a backbone during fine-tuning.
Specifically, the weights of both vision and text encoders are linearly interpolated between CLIP and the fine-tuned model as $\theta_w = (1-w) \cdot \theta_{\text{CLIP}} + w \cdot \theta_{\text{fine-tuned}}$.
TC-CLIP consistently outperforms others across all datasets, showing its superior generalization ability.

\vspace{.5em}
\noindent\textbf{Few-shot action recognition.}
We verify the learning capacity of our method under a challenging few-shot scenario.
In Table~\ref{p1_few_shot}, models are directly fine-tuned from CLIP on each dataset using $K$-shot samples, where $K$ is 2, 4, 8, and 16.
TC-CLIP achieves the best performance with large margins from ViFi-CLIP~\cite{vificlip}.

\vspace{.5em}
\noindent\textbf{Base-to-novel generalization.}
Similarly, models are directly fine-tuned from CLIP using the base classes of each dataset and evaluated for both base and novel classes.
Table~\ref{tab:base2novel} reports top-1 accuracies on the base and novel classes with their harmonic mean (HM).
TC-CLIP performs the best on the novel classes and HM across all datasets, especially showing  solid results on the SSv2 dataset.

\vspace{.5em}
\noindent\textbf{Fully-supervised action recognition.}
Table~\ref{p1_fullysupervised} shows performance comparison results under the fully-supervised setting, where 
the models are trained and evaluated both on the K-400 dataset.
TC-CLIP achieves top-1 accuracy of 85.2\% in the validation split, improving 1.3\%p over our baseline ViFi-CLIP~\cite{vificlip}.

\vspace{.5em}
\noindent\textbf{Computational cost.}
Table~\ref{tab:computation_cost_comparison} compares the computational cost with the average accuracy of all tasks.
We introduce a lightweight implementation of TC-CLIP (denoted by \S), where TC is only applied to the 4$^\text{th}$, 8$^\text{th}$, and 12$^\text{th}$ layers of the vision encoder.
Despite its reasonable cost, it performs best across all protocols by significant margins.
In particular, compared to Open-VCLIP~\cite{openvclip}, this lightweight version improves accuracy by 0.6\%p and 2.1\%p in the zero-shot and base-to-novel tasks, respectively, while maintaining 17.2\% higher throughput.

\begin{table*}[t]
\caption{\textbf{Component-wise ablations on the zero-shot setting.} $\Delta$ denotes the average top-1 accuracy gain over baseline.
}
\label{tab:ablations_component_wise}
\centering
\setlength{\tabcolsep}{2.8pt}
\vspace{-.5em}
\hspace{-2mm}
\scalebox{0.8}{
\begin{tabular}{lcccl|cccl} 
\toprule
& \multicolumn{4}{c}{Without weight-space ensembling} & \multicolumn{4}{c}{With weight-space ensembling} \\
\cmidrule(lr){2-9}
Case & HMDB-51 & UCF-101 & K-600 & All ($\Delta$) & HMDB-51 & UCF-101 & K-600 & All ($\Delta$)\\ 
\midrule
Baseline & 52.3\,$\pm$\,0.2 & 78.9\,$\pm$\,1.1 & 70.7\,$\pm$\,0.8 & 67.3 & 52.2\,$\pm$\,0.7 & 81.0\,$\pm$\,0.9 & 73.9\,$\pm$\,0.5 & 69.0\\
\midrule
(a) +TC & 53.6\,$\pm$\,0.2 & 78.6\,$\pm$\,1.0 & 71.8\,$\pm$\,0.7 & 68.0\,(\textcolor{darkergreen}{+0.7}) & 54.3\,$\pm$\,0.6 & 81.9\,$\pm$\,1.0 & 75.5\,$\pm$\,1.0 & 70.6\,(\textcolor{darkergreen}{+1.6})\\
(b) +VP & 53.2\,$\pm$\,0.8  & 80.5\,$\pm$\,0.7 & 71.6\,$\pm$\,0.9 & 68.4\,(\textcolor{darkergreen}{+1.1}) & 53.4\,$\pm$\,0.8 & 82.0\,$\pm$\,0.9 & 74.7\,$\pm$\,0.7 & 70.0\,(\textcolor{darkergreen}{+1.0})\\
\rowcolor{Gray!30}
(c) +TC+VP & 53.7\,$\pm$\,0.7 & 80.4\,$\pm$\,0.9 & 72.7\,$\pm$\,0.5 & 68.9\,(\textcolor{darkergreen}{+1.6}) & 54.2\,$\pm$\,1.1 & 82.9\,$\pm$\,0.9 & 75.8\,$\pm$\,0.4 & 71.0\,(\textcolor{darkergreen}{+2.0})\\
\bottomrule
\end{tabular}
}
\vspace{-1em}
\end{table*}

\begin{table*}[t]
\centering
\caption{\textbf{Effect of TC with various token aggregation strategies.}
TC consistently outperforms the frame-wise attention baseline across several different token selection and merging methods.
$K$-shot action recognition results are reported with the top-1 accuracy averaged over $K = 2, 4, 8, 16$.
Default settings are marked in \colorbox{baselinecolor}{gray}.
}
\label{tab:abl_aggregation}
\vspace{-1em}
\begin{minipage}{0.45\linewidth}{
\label{tab:abl_seed_token_selection}
\subcaption{{Seed token selection strategy.}}
\vspace{-2.5em}
\begin{center}
\setlength{\tabcolsep}{1pt}
\scalebox{0.73}{
\begin{tabular}{lccccl}
\toprule
Case & HMDB & UCF & SSv2 & All & ($\Delta$) \\
\midrule
Baseline & 62.6 & 89.2 & 8.7 & 53.5 \\
\midrule
No selection & 62.8 & 89.8 & 9.7 & 54.1 & (\textcolor{darkergreen}{+0.6}) \\
\midrule
Head-wise key norm & 62.3 & 89.8 & 9.8 & 54.0 & (\textcolor{darkergreen}{+0.5})\\
Averaged key norm & 62.5 & 89.4 & 9.3 & 53.7 & (\textcolor{darkergreen}{+0.2})\\
Head-wise CLS attn. & 63.4 & 89.9 & 9.7 & 54.3 & (\textcolor{darkergreen}{+0.8})\\
\rowcolor{Gray!30}
Averaged CLS attn. & 63.4 & 90.2 & \textbf{9.9} & \textbf{54.5} & (\textcolor{darkergreen}{+1.0})\\
Patch saliency~\cite{vidtldr} & 62.9 & \textbf{90.3} & 9.6 & 54.2 & (\textcolor{darkergreen}{+0.7})\\
ATS~\cite{ats} & \textbf{63.5} & \textbf{90.3} & 9.8 & \textbf{54.5} & (\textcolor{darkergreen}{+1.0})\\
\bottomrule
\end{tabular}
}
\end{center}}\end{minipage}
\noindent\hfill
\begin{minipage}{0.54\linewidth}{
\subcaption{{Context token summarization strategy.}}
\vspace{-2.5em}
\begin{center}
\setlength{\tabcolsep}{1pt}
\scalebox{0.73}{
\begin{tabular}{lccccl}
\toprule
Case & HMDB & UCF & SSv2 & All & ($\Delta$) \\
\midrule
Baseline & 62.6 & 89.2 & 8.7 & 53.5 \\
\midrule
No merge &57.2 & 85.6& 7.7 & 50.2 & (\textcolor{red}{$-3.3$}) \\
Random merge & 58.8 & 87.1 & 7.5 & 51.2 & (\textcolor{red}{$-2.3$}) \\
\midrule
K-means~\cite{kmeans} & 62.1 & 89.7& 9.0 & 53.6 & (\textcolor{darkergreen}{+0.1}) \\
DPC-KNN~\cite{dpcknn}  & 63.3 & \textbf{90.2} & 9.8 & 54.4 & (\textcolor{darkergreen}{+0.9})  \\
\rowcolor{Gray!30}
Bipartite soft matching~\cite{bipartite, tome} & \textbf{63.4} & \textbf{90.2} & \textbf{9.9}  & 54.5 & (\textcolor{darkergreen}{+1.0}) \\
Bipartite w/ attention weights & 62.9 & 89.8 & \textbf{9.9} & 54.2 & (\textcolor{darkergreen}{+0.7}) \\
Bipartite w/ saliency weights~\cite{vidtldr} & 62.4 & 89.9 & 9.6 & 54.0 & (\textcolor{darkergreen}{+0.5}) \\
\bottomrule
\end{tabular}
}
\end{center}}\end{minipage}
\end{table*}

\begin{table*}[t]
\caption{\textbf{TC design ablation.}
We report $K$-shot training results where the top-1 accuracy in each dataset is averaged over $K=2, 4, 8, 16$. Bias is defined in Eq.~\eqref{eq:learnable_bias}.}
\label{tab:ablations_tc}
\vspace{-1em}
\centering
\begin{minipage}{.47\linewidth}
\centering
\begin{subtable}[t]{\textwidth}
\centering
\caption{Positional embedding design.}
\vspace{-.5em}
\tabcolsep=1pt
\scalebox{0.75}{
\begin{tabular}{lcccc}
\toprule
Case & HMDB & UCF & SSv2 & All \\
\midrule
Spatial embedding &62.9 & 90.0 & 9.8 & 54.2 \\
Joint space-time embedding &63.2 & \textbf{90.2} & 9.8 & 54.4 \\
\rowcolor{Gray!30}
Spatial embedding + Bias & \textbf{63.4} & \textbf{90.2}& \textbf{9.9} & \textbf{54.5} \\
Joint embedding + Bias & 62.9 & \textbf{90.2}& 9.8 & 54.3 \\
\bottomrule
\\
\end{tabular}
}
\end{subtable}
\end{minipage}
\hfill\noindent
\begin{minipage}{.25\linewidth}
\centering
\begin{subtable}[t]{\textwidth}
\centering
\caption{Seed token ratio $\alpha$.}
\vspace{-.5em}
\tabcolsep=1pt
\scalebox{0.75}{
\begin{tabular}{ccccc}
\toprule
$\alpha$ & HMDB & UCF & SSv2 & All \\
\midrule
0.2 & 62.6 & 90.1 & 9.8 & 54.2 \\
\rowcolor{Gray!30}
    0.3 & \textbf{63.4} & 90.2 &\textbf{9.9} & 54.5 \\
    0.4 & 63.2 & \textbf{90.4} & 9.8 & 54.5 \\
    0.5 & 63.3 & 90.3 & 9.8 & 54.5 \\
    0.6 & 63.1 & 90.2 & 9.8 & 54.4 \\
\bottomrule
\end{tabular}
}
\end{subtable}
\end{minipage}
\hfill\noindent
\begin{minipage}{.25\linewidth}
\centering
\begin{subtable}[t]{\textwidth}
\centering
\caption{Context token $k$.}
\vspace{-.5em}
\tabcolsep=1pt
\scalebox{0.75}{
\begin{tabular}{ccccc}
\toprule
$k$ & HMDB & UCF & SSv2 & All \\
\midrule
16 & 63.1 & 89.3 & 9.1 & 53.8 \\
    32 & 63.6 & 89.9 & 9.4 & 54.3 \\
    64 & \textbf{63.7} & 90.1 & 9.7 & \textbf{54.5} \\
\rowcolor{Gray!30}
96 & 63.4 & \textbf{90.2} & \textbf{9.9} & \textbf{54.5} \\
    128 & 62.8 & 90.1 & \textbf{9.9} & 54.3 \\
\bottomrule
\end{tabular}
}
\end{subtable}
\end{minipage}
\vspace{-.5em}
\end{table*}
\begin{table*}[t]
\setlength{\tabcolsep}{3pt}
\caption{\textbf{Text prompting design ablation on the zero-shot setting.} All the models are evaluated without the weight ensemble.
}
\label{tab:abl_prompting_design}
\centering
\setlength{\tabcolsep}{4pt}
\vspace{-.5em}
\hspace{-2mm}
\scalebox{0.8}{
\begin{tabular}{lccccl}
\toprule
Case & Use context tokens? & HMDB-51 & UCF-101 & K-600 & All ($\Delta$) \\
\midrule
Baseline &  & 52.3 $\pm$ 0.2 & 78.9 $\pm$ 1.1      &  70.7 $\pm$ 0.8 & 67.3 \\ 
\midrule
(a) Learnable prompt vectors & & 52.4 $\pm$ 0.4 & 78.4 $\pm$ 1.3      &  70.6 $\pm$ 0.7 & 67.1 (\red{$-$0.2}) \\ 
(b) Video-conditional prompting & & 53.2 $\pm$ 0.8 & 80.4 $\pm$ 0.7      &  71.6 $\pm$ 0.9 & 68.4 (\textcolor{darkergreen}{+1.1}) \\ 
\rowcolor{Gray!30}
(c) Video-conditional prompting & \checkmark & 53.7 $\pm$ 0.7 & 80.4 $\pm$ 0.9      &  72.7 $\pm$ 0.5 & 68.9 (\textcolor{darkergreen}{+1.6}) \\ 
(d) Vision-text late-fusion & \checkmark & 53.7 $\pm$ 0.7 & 79.0 $\pm$ 0.7      &  70.9 $\pm$ 0.6 & 67.9 (\textcolor{darkergreen}{+0.6}) \\ 
\bottomrule
\end{tabular}
}
\end{table*}

\subsection{Analysis and Discussion}\label{sec:analysis}

This section examines the design choices and impact of each component in our model: Temporal Contextualization (TC) and Video-conditional Prompting (VP).
We mainly adopt the zero- and few-shot settings and report the average of top-1 accuracy with $K=2, 4, 8, 16$ for the $K$-shot setup.
In addition to the analyses discussed in this subsection, we present more analyses and qualitative results in the supplementary document.

\vspace{.5em}
\noindent\textbf{Component-wise ablation.}
Table~\ref{tab:ablations_component_wise} shows the impact of TC and VP on our baseline in the zero-shot setting.
Integrating TC gives an average gain of 0.7\%p over the baseline and the gap increases to 1.6\%p after adopting WE; WE is more favorable to our approach than the baseline.
Adopting VP also leads to a substantial gain of 1.1\%p, highlighting its own contribution.
When both VP and TC are applied to the baseline, an average improvement goes up to 1.6\%p, which finally leads to 2.0\%p gain after applying WE.

\vspace{.5em}
\noindent\textbf{Token aggregation strategies.}
In Table~\ref{tab:abl_aggregation}, we verify the effectiveness of TC across diverse token aggregation methods.
Experiments are conducted on the few-shot setting using the baseline model with TC.
(a) While TC still works well without token selection, we observe that collecting informative seed tokens based on token importance, such as attention or saliency scores, improves the quality of encoded tokens by suppressing the background.
(b) Directly using the seed tokens without merging reduces performance due to the extrapolation issue.
The degradation with random merging also highlights the requirement of token clustering based on relevance.
Finally, consistent gains from various token merging approaches verify the robustness of TC regardless of algorithms.

\vspace{.5em}
\noindent\textbf{Positional embedding.}
Table~\ref{tab:ablations_tc}(a) shows that using the proposed learnable bias (Eq.~\eqref{eq:learnable_bias}) with spatial positional embedding yields the best result.
We conjecture that the bias effectively consolidates the local frame-level information and global video-level information in a layer-wise and head-wise manner.

\vspace{.5em}
\noindent\textbf{Number of seed and context tokens.}
While TC is not sensitive to the choice of $\alpha$, as shown in Table~\ref{tab:ablations_tc}(b), we picked $\alpha=0.3$ as our default value, \ie, using 30\% of total tokens as seed tokens.
In Table~\ref{tab:ablations_tc}(c), the context token number $k$ is chosen to set a modest amount of merging degree.

\vspace{.5em}
\noindent\textbf{Text prompting design.}
In Table~\ref{tab:abl_prompting_design}, we observe that (a) a na\"ive integration of learnable prompt vectors without video instance conditioning is not particularly helpful for the zero-shot transferability, rather decreasing the average accuracy.
In contrast, (b) employing VP design with \classtoken~tokens consistently improves the accuracy across all datasets, and (c) using context tokens further enhances the performance, resulting in a 1.6\%p gain.
We also compare VP with (d) vision-text late-fusion design, \ie, the cross-attention of context tokens and the final representation of the text embedding. This design performs worse in UCF-101 and K-600 datasets than our VP, verifying the effectiveness of our design choice.

\begin{figure}[t]
\begin{minipage}{1.0\textwidth}
    \centering
    \begin{minipage}{0.49\textwidth}
    \centering
     \includegraphics[height=4.3cm, trim=0 2mm 0 0, clip]{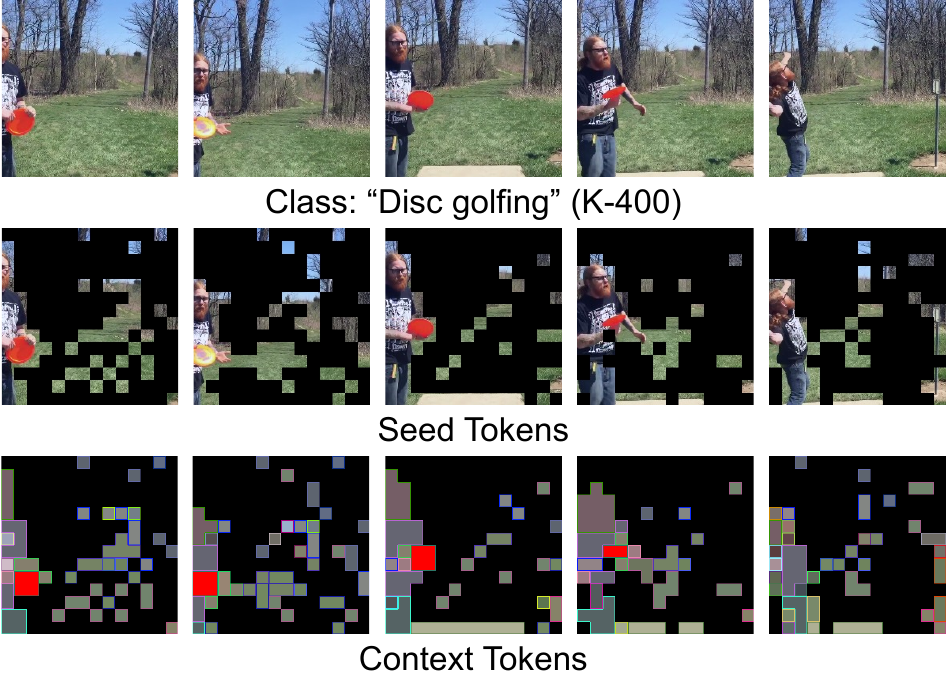}
     \vspace{-2em}
    \captionof{figure}{\textbf{Context token visualization.}
    TC-CLIP selects the informative seed tokens and summarizes them into context tokens across frames.
    The disc (\textcolor{red}{red}) is merged into one token over the video.
    }\label{fig:context_token_visualization}
    \end{minipage}
    \noindent \hfill
    \begin{minipage}{0.49\textwidth}
    \centering
     \includegraphics[height=4.3cm, trim=3mm 2mm 3mm 0, clip]{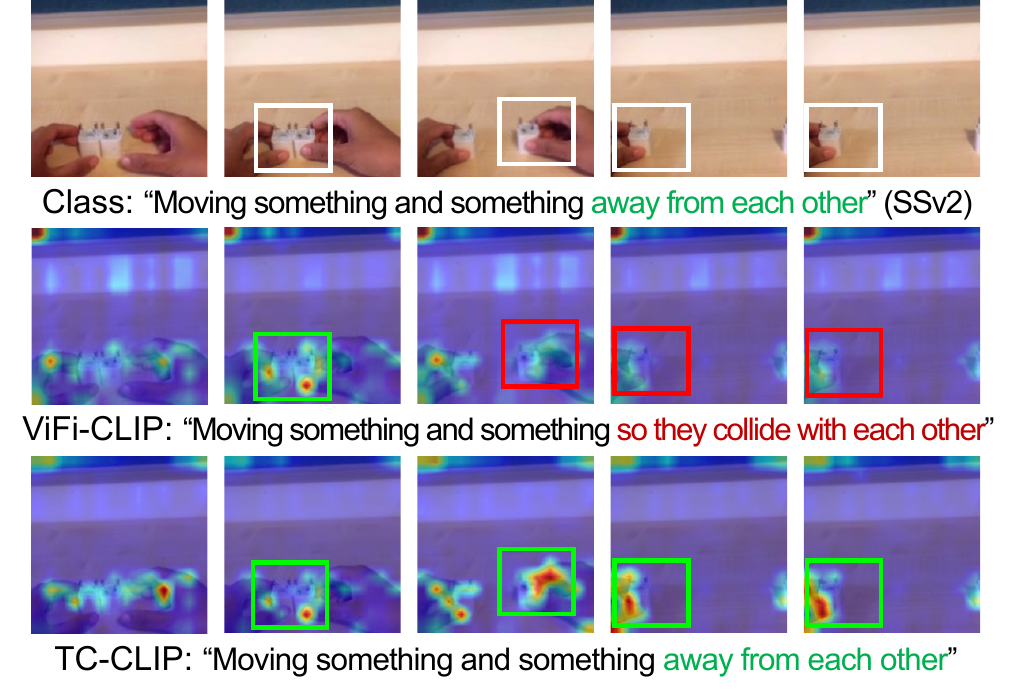}
     \vspace{-2em}
    \captionof{figure}{\textbf{Attention visualization.} While ViFi-CLIP fails to attend to the hands moving away and misinterprets the action as colliding, TC-CLIP correctly predicts by exploiting temporal consistency.}\label{fig:attention_visualization}
    \end{minipage}
\end{minipage}
\vspace{-1em}
\end{figure}

\vspace{.5em}
\noindent\textbf{Context token visualization.}
Fig.~\ref{fig:context_token_visualization} visualizes the seed tokens and context tokens from the last layer of the vision encoder in TC-CLIP.
In this video, the informative regions regarding the action of \textit{disc golfing} in each frame, including the person and the disc, are selected as seed tokens.
To visualize each context token, we colorize its corresponding source token positions using the average color of the input video patches of that region.
Note that a single context token (highlighted in \textcolor{red}{red}) successfully tracks the disc across multiple frames.

\vspace{.5em}
\noindent\textbf{Attention visualization.}
Fig.~\ref{fig:attention_visualization} visualizes the attention map of ViFi-CLIP~\cite{vificlip} and TC-CLIP on the SSv2 dataset.
In this video, where two hands grab objects and then move away, ViFi-CLIP~\cite{vificlip} fails to attend to the hands from the middle of the sequence and misinterprets the action as \textit{colliding with each other}.
In contrast, TC-CLIP considers the temporal context across the sequence by its design, and thus consistently attends to the hands throughout the entire video and correctly predicts the action as \textit{moving away from each other}.
\section{Related Work}\label{sec:related}

\noindent\textbf{Token aggregation.}
Recent studies on token aggregation aim to reduce the number of tokens given to image Transformers~\cite{dynamicvit, spvit, lit, evit, evovit, avit, ats, bat, tokenpooling, tcformer, tome, dtm} and video Transformers~\cite{tokenlearner, stts, testa, vidtome, sta, vidtldr} for efficient inference.
While some of these approaches train additional networks for token selection~\cite{dynamicvit, spvit, stts} or fusion~\cite{lit, tokenlearner}, we focus on parameter-free approaches, categorized into token pruning and merging.
Pruning-based methods~\cite{evit, evovit, avit, ats} eliminate uninformative tokens by measuring their importance using a metric such as a self-attention score, whereas merging-based methods combine tokens with large semantic similarity into single units using clustering algorithms such as $k$-means~\cite{tokenpooling}, DPC-KNN~\cite{tcformer}, and bipartite soft matching~\cite{tome, testa, vidtome}.
To minimize information loss, several studies~\cite{bat, sta, vidtldr} consider both token importance and similarities as aggregation criteria.
While our primary goal is not to improve efficiency, we employ both pruning and merging techniques to connect relevant tokens and summarize essential contexts within videos.
Although there are semantic segmentation studies~\cite{groupvit, ovsegmentor, segclip} that link relevant spatial data, they rely on learnable cluster centers with slot- or cross-attention blocks, and thus differ from our approach.

\vspace{.5em}
\noindent\textbf{Prompt learning.}
Several studies on prompt learning~\cite{vpt, coop, cocoop, maple, vificlip, vitaclip, a5} transfer VLMs to downstream tasks by optimizing a discrete set of prompt vectors.
In video recognition, \cite{a5} has introduced text prompt tuning, while ViFi-CLIP~\cite{vificlip} and Vita-CLIP~\cite{vitaclip} perform prompting in both vision and text branches.
However, these prompt vectors are separately optimized and not shared across the modalities.
In image recognition, Co-CoOp~\cite{cocoop} performs an instance-conditional prompt tuning by explicitly conditioning text prompts on the \classtoken\\ tokens from image instances.
MaPLe~\cite{maple} learns multi-modal prompting by sharing layerwise context prompts for both branches.
Unlikely, we generate video-conditional prompts by utilizing contextualized tokens as vision inputs and injecting summarized video information into text prompt vectors.

\section{Conclusion}\label{sec:conclusion}
We have introduced TC-CLIP, a novel video understanding paradigm that leverages holistic video information within the encoding process.
Unlike prior approaches that access only a limited range of tokens, the proposed temporal contextualization summarizes informative tokens from the entire video and utilizes them for attention operations.
While these tokens are employed to infuse temporal information on the vision side, they also serve as a source for video-conditional text prompts, thus enhancing the instance-wise context on the text side.
Extensive experiments and analyses on diverse benchmarks and evaluation protocols demonstrate the superiority of TC-CLIP and justify its design choices.

\section*{Acknowledgements}
Experiments are based on the NAVER Smart Machine Learning NSML~\cite{kim2018nsml} platform. 
This research was partly supported by the Bio \& Medical Technology Development Program of the National Research Foundation (NRF) (No. 2021M3A9E4080782) and the IITP grants [No. RS-2021-II212068; No. RS-2021-II211343] funded by the Korean government (MSIT).

\bibliographystyle{splncs04}
\bibliography{main}

\clearpage
\title{Leveraging Temporal Contextualization for\\Video Action Recognition\\--------Supplementary Material--------}
\titlerunning{TC-CLIP}
\author{ }
\authorrunning{M.~Kim et al.}
\institute{ }
\maketitle
\appendix
\setcounter{table}{10}
\setcounter{figure}{7}

\noindent
We provide additional experimental analyses and details in the following order:
\begin{itemize}
    \item Appendix~\ref{sec:sup_finetune_pretrained}: Fine-tuning with the Kinetics-400 pretrained model
    \item Appendix~\ref{sec:sup_abl_vp}: More ablation study on VP
    \item Appendix~\ref{sec:sup_vit_l}: Scalability with ViT-L/14
    \item Appendix~\ref{sec:sup_temporal_subset_analysis}: Temporal subset analysis
    \item Appendix~\ref{sec:sup_abl_we}: Impact of weight-space ensembling
    \item Appendix~\ref{sec:sup_visualization}: More visualizations of context tokens and attentions
    \item Appendix~\ref{sec:sup_details}: Datasets and implementation details
\end{itemize}

\section{Fine-tuning with the Kinetics-400 Pretrained Model}\label{sec:sup_finetune_pretrained}

\vspace{-2em}
\begin{table*}[h]
\centering
\caption{\textbf{Comparison with state-of-the-arts on few-shot action recognition using Kinetics-400 pretrained model.}
All the models are first pretrained on Kinetics-400 and subsequently fine-tuned on each dataset.
}
\label{tab:few_shot_pretrained}
\vspace{-.5em}
\hspace{-3mm}
\setlength{\tabcolsep}{2.8pt}
\scalebox{0.8}{
\begin{tabular}{lcccc|cccc|cccc|c}
  \toprule
    &\multicolumn{4}{c}{HMDB-51} & \multicolumn{4}{c}{UCF-101} & \multicolumn{4}{c}{SSv2} & All\\
  \cmidrule(lr){2-14}
  Method & $K$=$2$ & $K$=$4$ & $K$=$8$ & $K$=$16$ & $K$=$2$ & $K$=$4$ & $K$=$8$ & $K$=$16$ & $K$=$2$ & $K$=$4$ & $K$=$8$ & $K$=$16$ & ~Avg.~\\
  \midrule
  ActionCLIP~\cite{actionclip}  & 54.3	&56.2	&59.3	&66.1	&76.7	&80.4	&87.6	&91.8	&4.8	&6.9	&9.1	&12.3 & 50.5\\
  A5~\cite{a5} & 46.7&	50.4&	61.3&	65.8&	76.3&	84.4&	90.7&	93.0&	4.5&	  6.7&	7.2	& 9.5 & 49.7\\
  X-CLIP~\cite{xclip} &  60.5&	 66.8&	 69.3&	 71.7&	89.0&	 91.4&	 94.7&	 96.3&	 6.6	&  7.8&	 9.9	&  13.7 & 56.5\\
  ViFi-CLIP~\cite{vificlip}  & 63.0	& 65.1 &	69.6 &	72.0 &	91.0 &	93.7 &	95.0 &	96.4 &	6.7	& 7.9 &	10.2 &	13.5 & 57.0\\
\rowcolor{NavyBlue!10}
  \Ours~(Ours)   & \textbf{65.3} & \textbf{68.5} & \textbf{71.4} & \textbf{73.0} &\textbf{94.1} & \textbf{95.6} & \textbf{96.6} & \textbf{97.3} & \textbf{8.7} & \textbf{10.1} & \textbf{12.1} & \textbf{15.2} & \textbf{59.0} \\
\bottomrule                             
\end{tabular}
}
\end{table*}

\vspace{-3em}

\begin{table*}[h]
\centering
\setlength{\tabcolsep}{3pt}
\caption{\textbf{Comparison with state-of-the-arts on base-to-novel generalization using Kinetics-400 pretrained model.} 
All the models are first pretrained on Kinetics-400 and subsequently fine-tuned on each dataset.
}
\label{tab:base2novel_pretrained}
\vspace{-1em}
\hspace{-3mm}
\setlength{\tabcolsep}{4.5pt}
\scalebox{0.8}{
\begin{tabular}[t]{lccc|ccc|ccc|ccc}
\toprule
&\multicolumn{3}{c}{HMDB-51}  &\multicolumn{3}{c}{UCF-101}    & \multicolumn{3}{c}{SSv2} & \multicolumn{3}{c}{All (Avg.)} \\
\cmidrule(lr){2-13}
Method  & Base & Novel & HM         & Base & Novel & HM         & Base & Novel & HM   & Base & Novel & HM   \\
\midrule
ActionCLIP~\cite{actionclip}         & 69.0 & 57.2 & 62.6        & 85.6 & 75.3 & 80.1       &  8.1 &  8.7 &  8.4  &  54.2 &  47.1 &  50.4 \\
A5~\cite{a5}     & 70.4 & 51.7 & 59.6 & 95.8 & 71.0 & 81.6 & 12.9 & 5.7 & 7.9 & 59.7 & 42.8 & 49.9\\
X-CLIP~\cite{xclip}    & 75.8 & 52.0 & 61.7 & 95.4 & 74.0 & 83.4 & 14.2 & 11.0 & 12.4 & 61.8 & 45.7 & 52.5 \\
ViFi-CLIP~\cite{vificlip} & 77.1 & 54.9 & 64.1 & 95.9 & 74.1 & 83.6 & 15.8 & 11.5 & 13.3 & 62.9 & 46.8 & 53.7 \\
\rowcolor{NavyBlue!10}
\Ours~(Ours)  & \textbf{79.4} & \textbf{58.3} & \textbf{67.2} & \textbf{97.5} & \textbf{84.5} & \textbf{90.5} & \textbf{19.6} & \textbf{15.6} & \textbf{17.4} & \textbf{65.5} & \textbf{52.8} & \textbf{58.5}\\ 
\bottomrule
\end{tabular}
}
\end{table*}

\vspace{-1em}

\noindent
Tables \ref{tab:few_shot_pretrained} and \ref{tab:base2novel_pretrained} present the comparison results using the K-400 pretrained model on the few-shot and base-to-novel settings.
All the models are first pretrained on the Kinetics-400 dataset and subsequently fine-tuned on each dataset.
TC-CLIP demonstrates superior performance over all other methods by significant margins.
Particularly in the base-to-novel setup, TC-CLIP outperforms ViFi-CLIP~\cite{vificlip} with notable gaps of 3.5\%p, 6\%p, and 4.8\%p in the base, novel, and harmonic mean (HM) on average, respectively.

\section{More Ablation Study on VP}\label{sec:sup_abl_vp}

\begin{table*}[t]
\caption{\textbf{Video-conditional Prompting (VP) ablation.}
We report $K$-shot training results where the top-1 accuracy in each dataset is averaged over $K=2, 4, 8, 16$.
Default settings are marked in \colorbox{baselinecolor}{gray}.}
\label{tab:ablations_vp}
\vspace{-1em}
\centering
\begin{minipage}{.4\linewidth}
\centering
\begin{subtable}[t]{\textwidth}
\centering
\caption{Number of prompt vectors.}
\vspace{-.5em}
\tabcolsep=5pt
\scalebox{0.8}{
\begin{tabular}{ccccc}
\toprule
Case & HMDB & UCF & SSv2 & All \\
\midrule
2 & 62.0 & 90.1 & 9.4 & 53.8 \\
\rowcolor{Gray!30}
4 & \textbf{63.9} & \textbf{90.7} & \textbf{9.8} & \textbf{54.8} \\
8 & 63.7 & 90.4 & \textbf{9.8} & 54.7 \\
\bottomrule
\end{tabular}
}
\end{subtable}
\end{minipage}
\hfill\noindent
\begin{minipage}{.59\linewidth}
\centering
\begin{subtable}[t]{\textwidth}
\centering
\caption{Vision input token selection.}
\vspace{-.5em}
\tabcolsep=4pt
\scalebox{0.8}{
\begin{tabular}{lcccc}
\toprule
Case & HMDB & UCF & SSv2 & All \\
\midrule
CLS tokens from all frames & 63.3 & 90.3 & \textbf{9.8} & 54.4 \\
GAP tokens from all frames & 63.7 & 90.5 & \textbf{9.8} & 54.7 \\
\rowcolor{Gray!30}
Context tokens & \textbf{63.9} & \textbf{90.7} & \textbf{9.8} & \textbf{54.8} \\
\bottomrule
\end{tabular}
}
\end{subtable}
\end{minipage}
\\
\vspace{.5em}
\begin{minipage}{.4\linewidth}
\centering
\begin{subtable}[t]{\textwidth}
\centering
\caption{Input layer selection.}
\vspace{-.5em}
\tabcolsep=2pt
\scalebox{0.8}{
\begin{tabular}{cccccc}
\toprule
$L_\text{text}$ & $L_\text{vision}$ & HMDB & UCF & SSv2 & All \\
\midrule
1 & 1 & 62.3 & 89.7 & 9.1 & 53.7 \\
1 & 12 & 62.0 & 89.9 & 9.3 & 53.7 \\
6 & 6 & 62.8 & 89.5 & 9.7 & 54.0 \\
6 & 12 & 62.5 & 90.2 & 9.7 & 54.1 \\
\rowcolor{Gray!30}
12 & 12 & \textbf{63.9} & \textbf{90.7} & \textbf{9.8} & \textbf{54.8} \\
\bottomrule
\end{tabular}
}
\end{subtable}
\end{minipage}
\hfill\noindent
\begin{minipage}{.59\linewidth}
\centering
\begin{subtable}[t]{\textwidth}
\centering
\caption{Layer and prompt initialization.}
\vspace{-.5em}
\tabcolsep=3pt
\scalebox{0.8}{
\begin{tabular}{llcccc}
\toprule
Layer init. & Prompt init. & HMDB & UCF & SSv2 & All \\
\midrule
\rowcolor{Gray!30}
CLIP weight & ``\texttt{a photo of a}'' & \textbf{63.9} & \textbf{90.7} & 9.8 & \textbf{54.8} \\
Random init.& ``\texttt{a photo of a}'' & 63.5 & 90.5 & \textbf{9.9} & 54.6 \\
CLIP weight & Random init. & 62.5 & 90.2 & 9.5 & 54.1 \\
\bottomrule
\\
\\
\end{tabular}
}
\end{subtable}
\end{minipage}
\end{table*}
\begin{table}[t]
\centering
\caption{
\textbf{Computational cost--performance trade-off of VP design.} In the case of vision-text late-fusion design, class name embeddings are pre-computed. Models are evaluated on the zero-shot setting without the weight ensemble. Costs are measured using a single A6000 GPU.
}
\setlength{\tabcolsep}{3pt}
\vspace{-.5em}
\hspace{-2mm}
\scalebox{0.8}{
\begin{tabular}{l|c|ccc|cc} 
\toprule
    & K-600 & \multicolumn{3}{c|}{GFLOPs} & \multicolumn{2}{c}{Latency (s)} \\
Case & Top-1 & Vision & Text & Cross-modal & Training & Inference \\ 
\midrule
Vision-text late-fusion & 70.9 & 301 & 2.96 & 0.06 & 0.54 (1.00$\times$) & 0.042 (1.00$\times$)\\
\rowcolor{Gray!30}
Video-conditional Prompting (VP) & 72.7 (\textcolor{darkergreen}{+1.8}) & 301 & 2.96 & 0.06 & 0.58 (1.07$\times$) & 0.045 (1.07$\times$)\\
\bottomrule
\end{tabular}
}
\label{tab:abl_vp_efficiency}
\end{table}

Table~\ref{tab:ablations_vp} examines the design choice of VP in TC-CLIP on the few-shot setting.

\vspace{.3em}
\noindent\textbf{Number of prompt vectors.}
Increasing the number of prompt vectors does not necessarily improve performance. 4 prompt vectors are employed by default.

\vspace{.3em}
\noindent\textbf{Vision token selection.}
Using context tokens in VP yields better results than employing \classtoken~tokens or global average pooled (GAP) tokens from all frames.
This demonstrates that proper contextualization of vision features is essential to transfer the video information to the text side.

\vspace{.3em}
\noindent\textbf{Input layer selection.}
We vary the layer indices of the text and vision inputs $\{L_\text{text}, L_\text{vision}\}$ in the VP module $f_{\theta_\text{VP}}(\mathbf{p}^{L_\text{text}-1}, \mathbf{s}^{L_\text{vision}}_\text{proj})$.
We observe that conditional prompting at the early stage ($L_\text{text} = 1$) does not generalize well, regardless of the vision layer index.
The early-stage prompting design is hard to generalize in a full fine-tuning scenario, possibly because CLIP was initially trained in a vision-text late-alignment fashion.
Consequently, we choose the late-stage prompting by adopting the last layers for both modalities.

\vspace{.3em}
\noindent\textbf{Layer and prompt initialization.}
We initialize the VP module's weight using the weight from the last layer of the CLIP text encoder because random initialization often results in unstable training results in the few-shot scenario.
Similarly, it is beneficial to initialize the learnable prompt vectors using the prompt template ``\texttt{a photo of a}'' following several prompt tuning methods~\cite{cocoop, maple}.

\vspace{.3em}
\noindent\textbf{Computational cost analysis.}
Although VP requires the instance-conditional computation of text embeddings, the added cost is minor.
As in Table~\ref{tab:abl_vp_efficiency}, for a pair of video and text inputs, the GFLOPs required by the text encoder cost only about 1\% of those needed by the vision encoder.
Given these minimal text-related costs, VP adds only an extra 0.07$\times$ in latency compared to the vision-text late-fusion design using pre-computed text embeddings.
Considering the observed performance gain, this is an acceptable trade-off.

\begin{table*}[t]
\centering
\setlength{\tabcolsep}{4pt}
\caption{\textbf{Comparison with state-of-the-arts on zero-shot action recognition using ViT-L/14.} All the models are trained on Kinetics-400 and directly evaluated on other datasets. \textdagger~denotes that the results are reproduced with our implementation.
The best results are in \textbf{bold-faced} numbers, and the second-best ones are \underline{underlined}.
}
\label{tab:zero_shot_vit_l}
\vspace{-.5em}
\hspace{-2mm}
\tabcolsep=7pt
\scalebox{0.8}{
\begin{tabular}{lcccccc} 
\toprule
Method & WE & HMDB-51 & UCF-101 & K600 (Top-1) & K600 (Top-5) & All (Top-1) \\ 
\midrule
ViFi-CLIP~\cite{vificlip}$^\dagger$ & & 55.6 $\pm$ 0.5 & 86.1 $\pm$ 0.8 & 77.8 $\pm$ 0.9 & 95.6 $\pm$ 0.2 & 73.2 \\
\rowcolor{NavyBlue!10}
\Ours~(Ours) & & \textbf{56.1} $\pm$ 0.3 & \textbf{86.9} $\pm$ 0.9 & \textbf{80.1} $\pm$ 0.7 & \textbf{96.5} $\pm$ 0.1 & \textbf{74.4} \\
\midrule
ViFi-CLIP~\cite{vificlip}$^\dagger$ & \checkmark & 55.8 $\pm$ 0.7 & \underline{88.1} $\pm$ 1.3 & \underline{81.1} $\pm$ 0.7 & 96.7 $\pm$ 0.1 & 75.0 \\
Open-VCLIP~\cite{openvclip} & \checkmark & \textbf{59.0} $\pm$ 0.6 & 87.6 $\pm$ 1.2 & \underline{81.1} $\pm$ 0.8 & 96.3 $\pm$ 0.3 & \underline{75.9} \\
\rowcolor{NavyBlue!10}
\Ours~(Ours) & \checkmark & \underline{57.1} $\pm$ 0.7 & \textbf{88.9} $\pm$ 0.9 & \textbf{83.1} $\pm$ 0.7 & \textbf{97.3} $\pm$ 0.1 & \textbf{76.4} \\
\bottomrule
\end{tabular}
}
\end{table*}

\begin{table*}[t]
\centering
\setlength{\tabcolsep}{4pt}
\caption{\textbf{Temporal subset analysis} using the temporal subset~\cite{temporalsubset} on Kinetics-400 and SSv2.
Gains over ViFi-CLIP are indicated in \textcolor{darkergreen}{green}.
}
\label{tab:temporal_subset_analysis}
\vspace{-1em}
\hspace{-1mm}
\tabcolsep=12pt
\scalebox{0.8}{
\begin{tabular}[t]{lll|ll}
\toprule
&\multicolumn{2}{c}{K-400 fully-supervised}  &\multicolumn{2}{c}{SSv2 16-shot} \\
\cmidrule(lr){2-5}
Method  & All & Temporal & All & Temporal   \\
\midrule
ViFi-CLIP~\cite{vificlip} & 83.9 & 87.8 & 12.4 & 25.9 \\
\Ours~(Ours)  & 85.2 (\textcolor{darkergreen}{$+$1.3}) & 89.2 (\textcolor{darkergreen}{$+$1.4}) & 14.0  (\textcolor{darkergreen}{$+$1.6}) & 29.9  (\textcolor{darkergreen}{$+$4.0}) \\ 
\bottomrule
\end{tabular}
}
\end{table*}

\section{Scalability with ViT-L/14}\label{sec:sup_vit_l}
Table~\ref{tab:zero_shot_vit_l} shows the zero-shot performance comparison using CLIP ViT-L/14 as a backbone.
In the case of using WE, our model outperforms ViFi-CLIP~\cite{vificlip} and Open-VCLIP~\cite{openvclip} by 1.4\%p and 0.5\%p on average, respectively.

\section{Temporal Subset Analysis}\label{sec:sup_temporal_subset_analysis}
We adopt the temporal subset analysis suggested in~\cite{temporalsubset} to further analyze the temporal modeling ability of trained models.
The temporal subset consists of several action classes that require more temporal information to recognize them, \ie, the classes of videos that cannot be recognized by human annotators after randomly shuffling the frames.
As shown in Table~\ref{tab:temporal_subset_analysis}, TC-CLIP's gains over ViFi-CLIP~\cite{vificlip} on the temporal subsets are more substantial than the gains when evaluated on the full validation splits, demonstrating the superiority in handling temporal information.

\section{Impact of Weight-space Ensembling}\label{sec:sup_abl_we}

\begin{wrapfigure}{l}{0.5\textwidth}
  \vspace{-38pt}
  \begin{center}
    \includegraphics[width=0.48\textwidth]{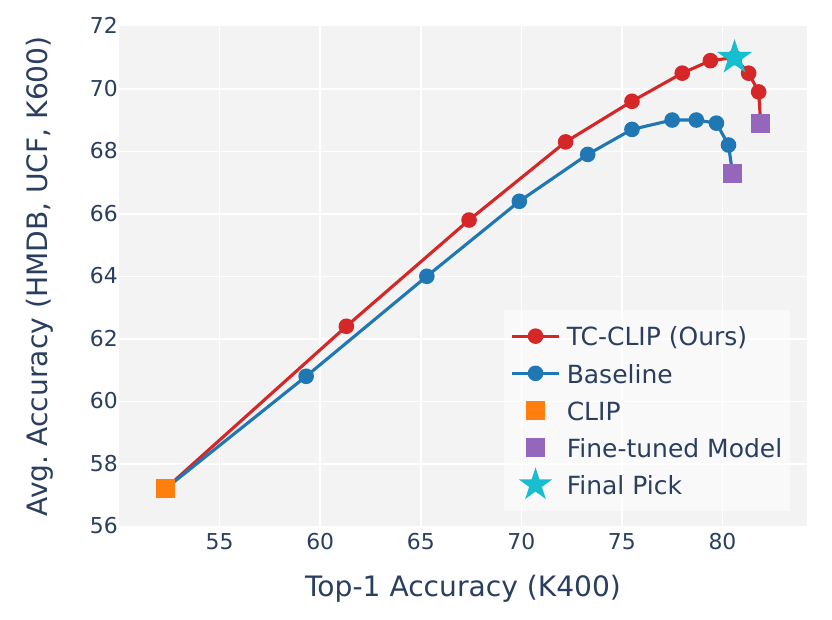}
  \end{center}
  \vspace{-18pt}
  \caption{\textbf{Weight averaging ablation.} 
    \label{fig:weight_ensemble_curve}}
  \vspace{-18pt}
\end{wrapfigure}
In Fig.~\ref{fig:weight_ensemble_curve}, we evaluate the effectiveness of weight ensembling by varying the ensemble ratio $w$ from 0 to 1 with a step size of 0.1.
Specifically, the backbone weights of both vision and text encoders are linearly interpolated between CLIP and fine-tuned model, \ie, $\theta_w = (1-w) \cdot \theta_{\text{CLIP}} + w \cdot \theta_{\text{fine-tuned}}$.
The $y$-axis shows the average accuracy on the zero-shot video datasets, and the $x$-axis means the accuracy on the fine-tuning dataset K-400.
Our model achieves a better trade-off than the baseline as our curve is always on top of the baseline's curve.
This demonstrates that our model takes more advantages from weight ensembling.
We choose $w=0.7$ as our final ensemble ratio.

\begin{figure}[p]
\centering
\includegraphics[width=\linewidth]{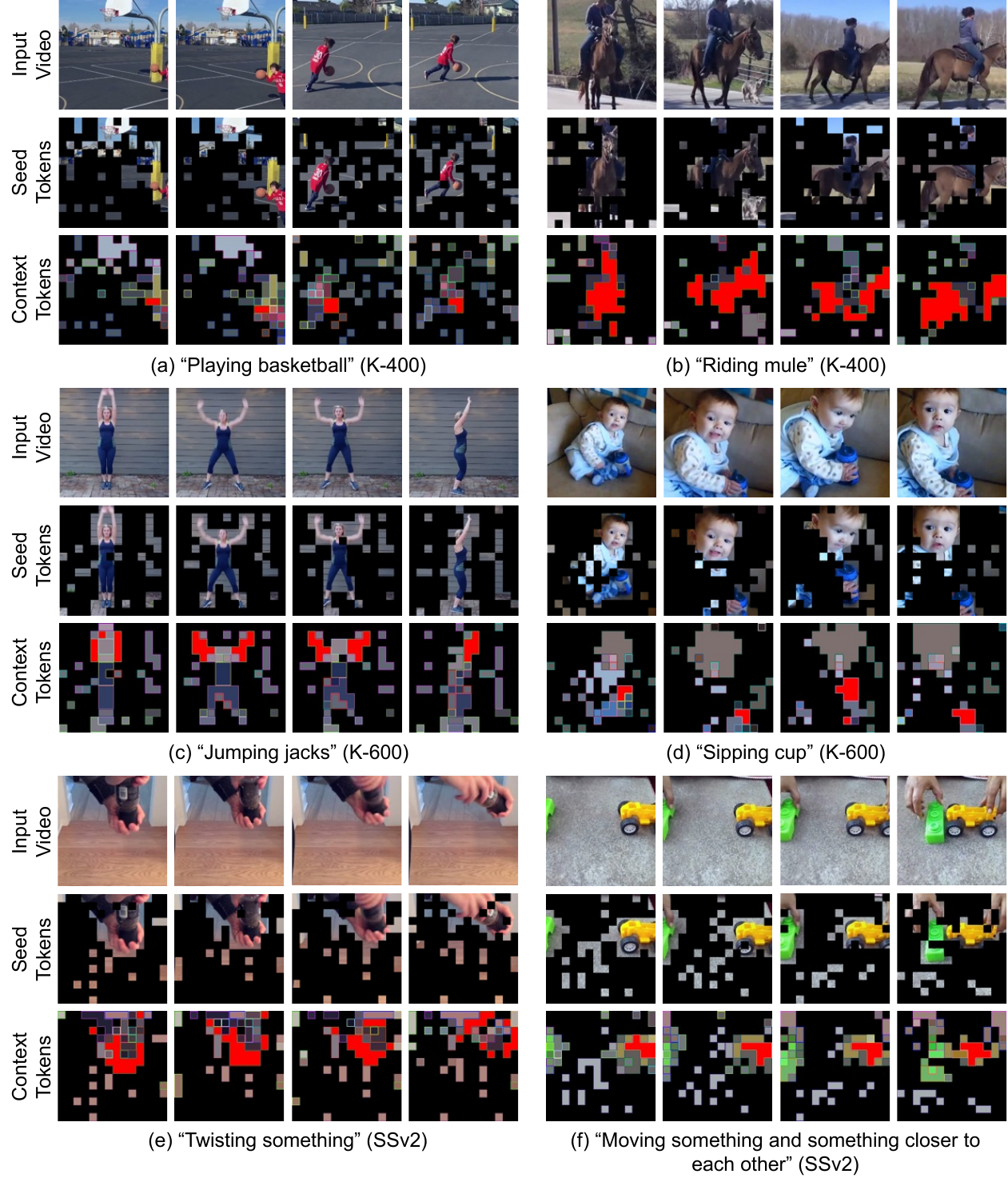}
\vspace{-1em}
\caption{
\textbf{Context token visualization} of TC-CLIP on Kinetics-400, Kinetics-600, and SSv2 datasets.
We visualize selected seed tokens and the resulting context tokens in the last layer of the vision encoder.
Patch tokens with the same inner and border color are summarized into one context token.
Regions highlighted in \textcolor{red}{red} represent a specific object or part grouped into a single context token throughout the video.
}
\label{fig:sup_merged_token_vis}
\vspace{-1em}
\end{figure}
\begin{figure}[p]
\centering
\includegraphics[width=\linewidth]{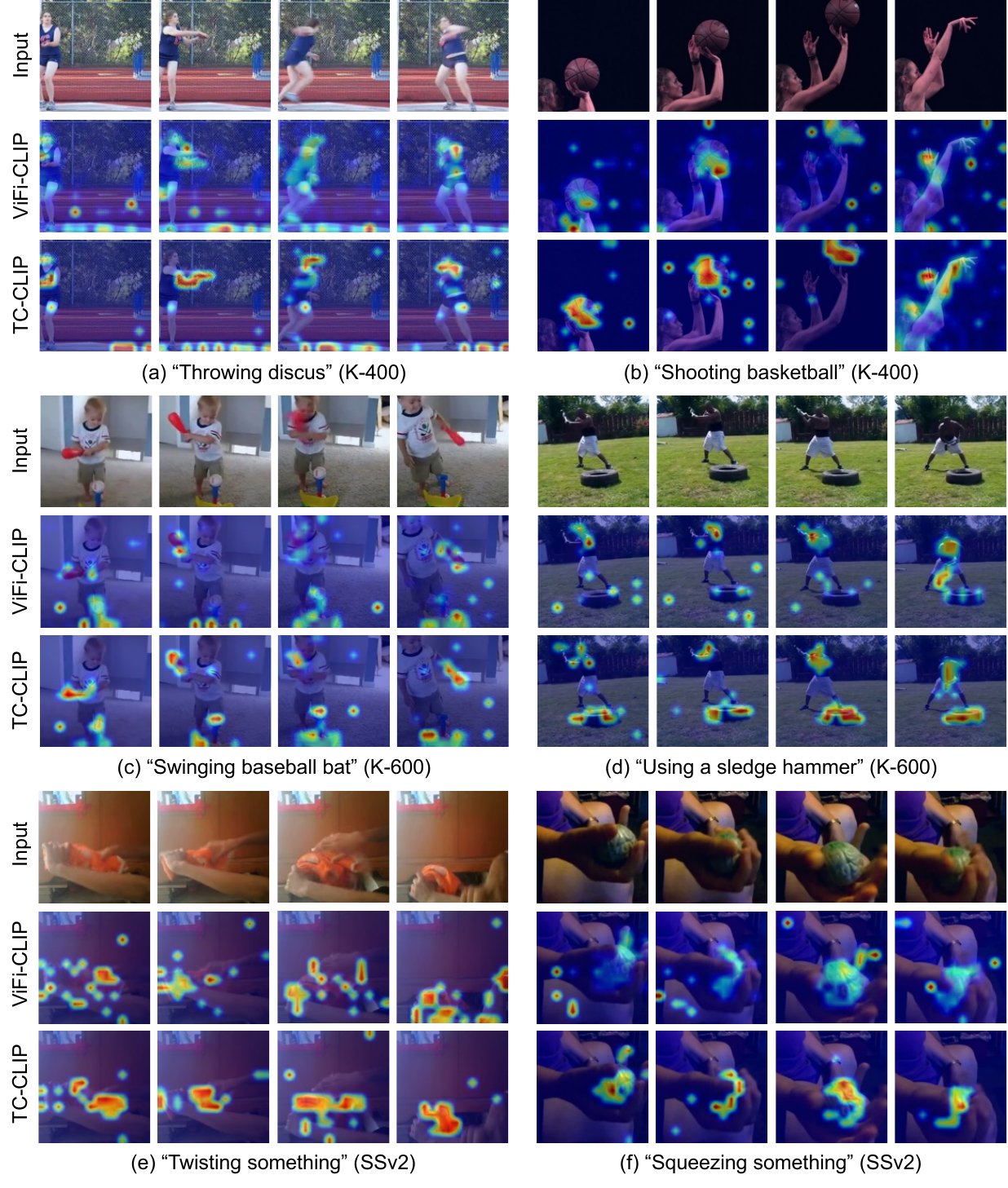}
\vspace{-1em}
\caption{
\textbf{Attention visualization} of TC-CLIP in comparison with ViFi-CLIP~\cite{vificlip} on Kinetics-400, Kinetics-600, and SSv2 datasets using \classtoken~token as a query in each frame.
(\textbf{a})--(\textbf{b}): TC-CLIP tends to focus more on fast-moving parts such as hands and arms.
(\textbf{c})--(\textbf{d}): While ViFi-CLIP dominantly attends to the most salient regions, TC-CLIP attends to multiple objects based on inter-object relationships relevant to the occurring actions.
(\textbf{e})--(\textbf{f}): TC-CLIP consistently attends to the main object with deformations throughout the video.
}
\label{fig:sup_attention_vis}
\vspace{-1em}
\end{figure}
\begin{figure}[p]
\centering
\begin{minipage}{\linewidth}
\includegraphics[width=\linewidth]{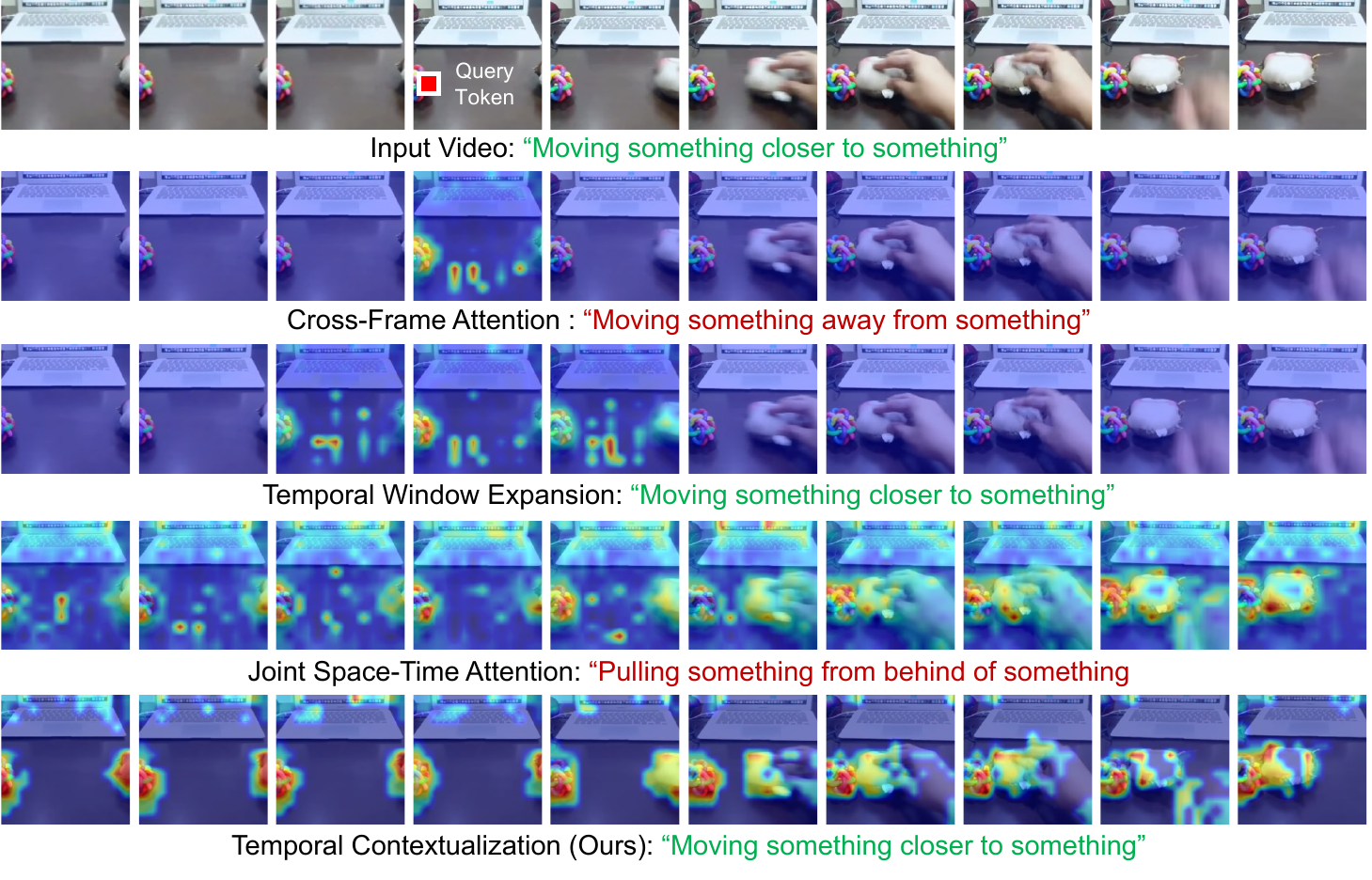}
\vspace{1em}
\end{minipage}
\begin{minipage}{\linewidth}
\vspace{-1em}
\includegraphics[width=\linewidth]{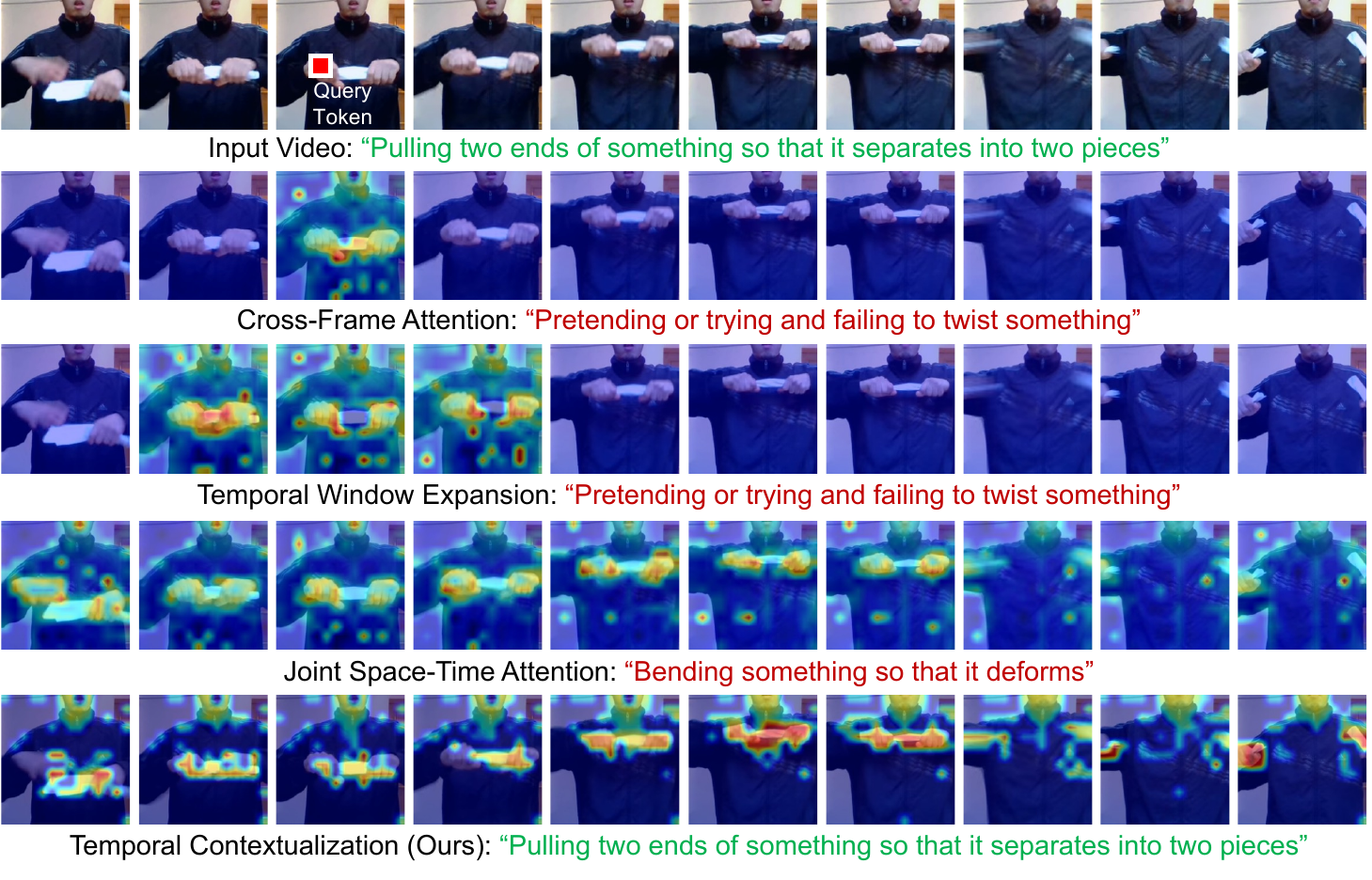}
\end{minipage}
\caption{
\textbf{Attention visualization} of TC-CLIP in comparison with various temporal information learning approaches on SSv2 dataset.
We visualize the attention map in the last vision encoder layer using a ball (top) and a hand (bottom) as a query (denoted with \textcolor{red}{red} boxes).
To visualize the attention map from TC, we assign attention values of context tokens to their corresponding source patch token positions.
Unlike other approaches, our method successfully highlights informative regions globally over frames.
}
\label{fig:sup_query_attention_vis}
\vspace{-1em}
\end{figure}

\section{More Visualizations of Context Tokens and Attentions}\label{sec:sup_visualization}

\noindent\textbf{Context token visualization.}
Fig.~\ref{fig:sup_merged_token_vis} visualizes the seed tokens and context tokens from the last layer of the vision encoder in TC-CLIP.
The seed tokens mainly consist of patch tokens from the most informative regions in each frame, often corresponding to the foreground, such as a person, animals, hands, and objects.
To visualize each context token, we colorize its corresponding source token positions using the average color of the input image patches of that region.
It is noteworthy that a single context token (highlighted in \textcolor{red}{red}) successfully tracks and summarizes a specific object or part throughout the entire video.

\vspace{0.5em}
\noindent\textbf{Class token attention visualization.}
Fig.~\ref{fig:sup_attention_vis} visualizes the attention maps of TC-CLIP compared to ViFi-CLIP~\cite{vificlip} using \classtoken~token as a query in each frame.
As shown in Fig.~\ref{fig:sup_attention_vis}(a)--(b), during the action of throwing or shooting objects, TC-CLIP tends to focus more on dynamically moving parts such as hands and arms.
Furthermore, as in Fig.~\ref{fig:sup_attention_vis}(c)--(d), TC-CLIP highlights multiple objects simultaneously based on inter-object relationships.
During actions like ``swinging baseball bat,'' TC-CLIP focuses on both the bat and the baseball being struck, whereas ViFi-CLIP only highlights salient areas in individual frames.
Fig.~\ref{fig:sup_attention_vis}(e)--(f) also shows TC-CLIP's consistent attention towards objects with deformations across frames, which is more striking than ViFi-CLIP's.

\vspace{0.5em}
\noindent\textbf{Patch token attention visualization.}
Fig.~\ref{fig:sup_query_attention_vis} shows the attention maps of TC-CLIP compared to other temporal modeling approaches~\cite{xclip, vitaclip, openvclip} by using a patch token as a query.
To visualize the attention map from TC, we assign attention values of context tokens to their corresponding source patch token positions.
In both examples, the token interactions of cross-frame attention~\cite{xclip, vitaclip} and temporal window expansion~\cite{openvclip} cannot reach the frames far from the query position, although the main action actually happens in the latter part of videos.
The joint space-time attention model, on the other hand, is capable of global modeling but fails to focus on informative regions.
In contrast, TC-CLIP consistently highlights the regions relevant to the query positions (\eg, hands and grabbed objects) throughout the video, leading to more accurate predictions.

\section{Experimental Setup Details}\label{sec:sup_details}

\subsection{Dataset Details}
We conduct experiments over 5 action recognition benchmarks: Kinetics-400~\cite{k400} \& 600~\cite{k600}, HMDB-51~\cite{hmdb51}, UCF-101~\cite{ucf101}, and Something-Something v2 (SSv2)~\cite{ssv2}.

\vspace{.3em}
\noindent \textbf{Kinetics-400}~\cite{k400} is a large-scale action recognition dataset with a total of 400 action classes, where its video clips are collected from YouTube and last for about 10 seconds. It contains around 240k training videos and 20k validation videos.

\vspace{.3em}
\noindent \textbf{Kinetics-600}~\cite{k600} is an extension of Kinetics-400 with approximately 480k video clips covering 600 action categories. The videos are divided into 390k for training, 30k for validation, and 60k for testing.
We mainly adopt the validation split for zero-shot evaluation.

\vspace{.3em}
\noindent \textbf{HMDB-51}~\cite{hmdb51} dataset includes 6,869 clips divided into 51 action categories. There are three individual splits for training and validation.

\vspace{.3em}
\noindent \textbf{UCF-101}~\cite{ucf101} is an action recognition dataset collected from YouTube, including 13,320 video clips with 101 action categories. Similar to HMDB-51, the training and test videos have three splits.

\vspace{.3em}
\noindent \textbf{SSv2}~\cite{ssv2} is a challenging dataset with 174 fine-grained action classes, which are more temporally biased than the other datasets. The standard split consists of 168,913 training videos and 24,777 validation videos.

\subsection{Implementation Details}
During the bipartite soft matching~\cite{bipartite, tome}, we start with the seed tokens arranged based on the \classtoken~token attention values in each frame.
These tokens are then divided into two sets by alternating positions.
Subsequently, $r$ pairs of tokens with the highest cosine similarity are merged by averaging their features, and the remaining two sets are then concatenated back together.
We set $r$ to 100 in practice.
This process is repeated iteratively, employing a constant $r$ scheduling for every iteration with an exception in the final iteration to ensure that the number of final context tokens becomes $k$.

During the training, we sample 16 frames to form a video clip.
During the evaluation, two temporal clips with one spatial crop (2 $\times$ 1 view) per video are sampled to produce a prediction unless otherwise stated.
The learnable prompts are initialized with the prompt ``\texttt{a photo of a}'' following~\cite{cocoop, maple}, and the weight of the VP module is initialized with the weight from the last layer of the CLIP text encoder.
For training recipes, we follow~\cite{vificlip} for zero-shot, few-shot, and fully-supervised settings and follow~\cite{froster} for base-to-novel generalization.
By default, we use the AdamW optimizer with momentum betas of (0.9, 0.98) and a weight decay of 0.001.
The VP module's initial learning rate is $10\times$ larger than the base learning rate in each setting.
Training configurations and evaluation metrics in each protocol are specified below.

\vspace{.3em}
\noindent\textbf{Zero-shot action recognition.}
The models are trained on Kinetics-400 and evaluated on HMDB-51, UCF-101, and Kinetics-600 datasets.
For HMDB-51 and UCF-101, we report the average and standard deviation of top-1 accuracy across three official validation splits.
In the case of Kinetics-600, we apply the zero-shot evaluation protocol from~\cite{zsar}, which exploits 220 categories of Kinetics-600 that do not appear in Kinetics-400.
We use the three splits provided by~\cite{zsar}, each containing 160 categories.
The results include the average top-1 and top-5 accuracy and their respective standard deviations.
During the training, the base learning rate is set to $8 \times 10^{-6}$ and is decayed to $8 \times 10^{-8}$ following the cosine decay scheduler.
The batch size is 256, and the total number of epochs is 10, including 5 linear warmup epochs.

\vspace{.3em}
\noindent\textbf{Few-shot action recognition.}
We adopt the $K$-shot training splits from~\cite{vificlip} that randomly sampled $K=2, 4, 8, 16$ videos from each class on HMDB-51, UCF-101, and SSv2.
The models are evaluated using the first validation split of HMDB-51 and UCF-101 and the full validation split of SSv2.
The base learning rate is set to $2 \times 10^{-6}$ and is decayed to $2 \times 10^{-8}$.
The batch size is 64, and the total number of epochs is set to 50, starting with 5 linear warmup epochs.

\vspace{.3em}
\noindent\textbf{Base-to-novel generalization.}
We adopt the base and novel splits from~\cite{vificlip}.
The models are trained on a set of base (seen) classes in a few-shot manner and subsequently evaluated on a set of novel (unseen) classes for four datasets: Kinetics-400, HMDB-51, UCF-101, and SSv2.
Each dataset comprises three training splits containing randomly sampled 16 shots of base action categories.
We report the average accuracy over three splits.
For HMDB-51 and UCF-101, the training and validation consider only their first split, whereas, for Kinetics and SSv2, the models are evaluated on their full validation split.
The base learning rate is set to $3.33 \times 10^{-6}$ and is decayed to $3.33 \times 10^{-8}$.
The batch size is 64.
The number of epochs is 12, including 2 warmup epochs.

\vspace{.3em}
\noindent\textbf{Fully-supervised action recognition.}
The models are trained on Kinetics-400 and evaluated on its complete validation split.
The base learning rate is set to $2.2 \times 10^{-5}$ and is decayed to $2.2 \times 10^{-7}$ following the cosine decay scheduler.
The batch size is 512, and the total epochs is 30 epochs, including 5 linear warmup epochs.

\end{document}